\documentclass{article}

\usepackage[preprint]{neurips_2026}

\usepackage[utf8]{inputenc} 
\usepackage[T1]{fontenc}    
\usepackage{hyperref}       
\usepackage{url}            
\usepackage{booktabs}       
\usepackage{amsfonts}       
\usepackage{nicefrac}       
\usepackage{microtype}      
\usepackage{xcolor}         
\usepackage{algorithm}
\usepackage{algorithmic}
\usepackage{makecell}
\usepackage{array}

\usepackage{threeparttable}
\usepackage{adjustbox}
\usepackage{amsmath}
\usepackage{amsthm}
\usepackage{cleveref}
\usepackage{amssymb}
\usepackage{xcolor}
\usepackage{pifont}
\usepackage{enumitem}

\theoremstyle{plain}
\newtheorem{theorem}{Theorem}

\theoremstyle{remark}

\definecolor{LightSteelBlue1}{RGB}{202,225,255}
\definecolor{LightPink}{RGB}{245,191,210}
\definecolor{Moccasin}{RGB}{255, 228, 181}

\definecolor{LightSteelBlue1}{RGB}{202,225,255}
\definecolor{LightPink}{RGB}{245,191,210}
\definecolor{Moccasin}{RGB}{255, 228, 181}
\setenumerate[1]{itemsep=0pt,partopsep=0pt,parsep=\parskip,topsep=0pt}
\setitemize[1]{itemsep=0pt,partopsep=0pt,parsep=\parskip,topsep=0pt}
\setdescription{itemsep=0pt,partopsep=0pt,parsep=\parskip,topsep=0pt}

\newcommand{\cmark}{\textcolor{green!60!black}{\ding{51}}}
\newcommand{\xmark}{\textcolor{red}{\ding{55}}}
\crefname{appendix}{appendix}{appendices}
\Crefname{appendix}{Appendix}{Appendices}

\usepackage{booktabs}
\usepackage{graphicx}
\usepackage{subcaption}

\usepackage{wrapfig}

\title{ChunkFT: Byte-Streamed Optimization for Memory-Efficient Full Fine-Tuning}

\author{
  Yongkang Liu$^{1}$
  Zijing Wang$^{1}$
  Mengjie Zhao$^{1}$
  Ercong Nie$^{2}$
  Mingyang Wang$^{3,4}$ \\
  \textbf{Qian Li$^{5}$}\thanks{Corresponding authors.}\hspace{0.3em}
  \textbf{Feiliang Ren}$^{1}$
  \textbf{Shi Feng}$^{1}$
  \textbf{Daling Wang}$^{1}$
  \textbf{and Hinrich Schütze$^{3,4}$} \\
  $^{1}$Northeastern University, China; $^{2}$Shanghai Jiao Tong University, China \\
  $^{3}$CIS, LMU Munich, Germany;$^{4}$MCML, Germany;
  $^{5}$Shandong University, China;
}

\begin{document}

\maketitle

\begin{abstract}
This work presents \textsc{ChunkFT}, a memory-efficient fine-tuning framework that reformulates full-parameter fine-tuning around a dynamically activated working set. \textsc{ChunkFT} enables gradient computation for arbitrary sub-tensors without modifying the network architecture, providing an algorithmic foundation for optimizing arbitrary sub-networks while avoiding standard dense gradient computation. We provide a theoretical convergence analysis of \textsc{ChunkFT} in the deterministic setting. Empirically, we apply \textsc{ChunkFT} to fine-tune Llama 3-8B and Llama 3-70B using a single RTX 4090-24GB GPU and 2$\times$ H800-80GB GPUs, respectively. Full-parameter fine-tuning of a 7B model with a 1K input length requires only 13.72GB of GPU memory. The results demonstrate the effectiveness of \textsc{ChunkFT} in memory usage, running time, and optimization quality. Moreover, downstream evaluations on language understanding, mathematical reasoning, and MT-Bench show that \textsc{ChunkFT} consistently outperforms existing memory-efficient baselines. Notably, \textsc{ChunkFT} achieves performance comparable to, and in some cases exceeding, full-parameter fine-tuning. Our repository is on~\url{https://github.com/misonsky/chunk}.

\end{abstract}

\section{Introduction}

Full-parameter fine-tuning (FFT) of language models (LMs) has proven to be a successful paradigm in various downstream tasks~\cite{lewis2020bart,devlin2019bert,radford2019language}. However, a major obstacle in fine-tuning is the substantial memory required, which escalates as models increase in size and complexity, thereby limiting the scalability and accessibility for those with limited computational resources~\cite{lv2024full,dettmers2023qlora,chen2016training,rajbhandari2020zero}.

To mitigate the memory constraints, Parameter-Efficient Fine-Tuning (PEFT) has been introduced to address memory limitations by updating only a small subset of parameters, while still achieving results comparable to full-model fine-tuning~\cite{HuSWALWWC22,zaken2022bitfit,yu2025ssmlora,houlsby2019parameter,li2021prefix,lester2021power}. Although PEFT methods have yielded promising results, existing studies indicate that they exhibit performance gaps compared to FFT in areas such as complex reasoning tasks and robustness~\cite{liu2025look}.

In response to these challenges, MEFT (Memory-Efficient Fine-Tuning) presents a promising solution. MEFT typically reduces memory usage during FFT by improving optimizer design~\cite{lv2024full,lv2024adalomo,zhaogalore,zhu2025apollo} or optimizing the backpropagation (BP) process~\cite{luo2024badam,liu2024hift}. Optimizer-based methods typically optimize only gradients and optimizer states, without addressing activation storage; consequently, their overall memory savings remain limited~\cite{liu2024hift,lv2024full,lv2024adalomo}. BP process optimization methods generally adopt layer-wise scheduling, represented by HiFT~\cite{liu2024hift} and LISA~\cite{pan2024lisa}. These methods update only a subset of parameters at each step and achieve full-parameter updates through an iterative process, thereby reducing GPU memory usage for activations, gradients, and optimizer states simultaneously. Nevertheless, these methods suffer from two key shortcomings: \textbf{GPU memory fluctuations} and \textbf{wasteful computational overhead}.
In language models, layer sizes vary across embeddings, attention layers, and MLP blocks, causing the mutable-state footprint to fluctuate and leading to unstable memory peaks that severely limit GPU usability. 
Moreover, during backpropagation, these methods still propagate gradients through inactive layers, only to discard them afterward, resulting in a significant waste of computational resources.

Rather than binding the update granularity to layers, or preserving dense backward computation and then discarding most gradients afterward, we seek a method that 
(i) partitions trainable tensors 
at a finer granularity 
than individual layers
using their actual byte-level training cost, 
(ii) materializes gradients only on the active support during backward, and 
(iii) keeps optimizer residency, state placement, and activation re-materialization synchronized to that same support so that memory savings are accompanied by a stable temporal memory profile.

We pursue this perspective through \textsc{ChunkFT}, a memory-centric fine-tuning framework that activates only a scheduled subset of trainable parameter slices at each optimization step. Chunks are balanced by a byte budget rather than parameter count. 
We implement arbitrary sub-tensor gradient updates within the PyTorch framework. During BP, \textsc{ChunkFT} materializes gradients only for the active slices into auxiliary buffers. To accelerate training, we use asynchronous prefetching and offloading to overlap state transfers with computation. Fine-tuning is thereby transformed from a globally dense procedure into a locally active one, with memory residency concentrated on a rotating working set rather than the full parameter space.
This design yields both practical and conceptual benefits. Practically, it lowers peak device memory while preserving the original training objective and remaining compatible with standard transformer training pipelines. 
Furthermore, a theoretical analysis is provided to confirm the convergence of ChunkFT.

Our contributions are threefold: \textbf{(1)} We propose a memory-efficient fine-tuning framework \textsc{ChunkFT}, which reformulates full-parameter fine-tuning around a dynamically activated working set. We provide a convergence analysis for ChunkFT, demonstrating that ChunkFT’s update rule yields a convergent scheme. \textbf{(2)} We develop chunk-local training operators that enable gradient computation on arbitrary sub-tensors without modifying the original network architecture, making it broadly applicable to existing standard deep learning frameworks and transformer architectures. \textbf{(3)} We apply \textsc{ChunkFT} to fine-tune Llama3-8B and Llama3-70B models using \emph{a single} RTX 4090-24GB GPU and 2$\times$ H800-80GB GPUs, respectively. Extensive evaluations on language understanding, mathematical reasoning, and MT-Bench benchmarks show that \textsc{ChunkFT} consistently outperforms existing memory-efficient baselines while achieving performance comparable to full-parameter fine-tuning.
    

\begin{table*}[t]
\captionsetup{skip=2pt}
\centering
\caption{
Algorithm feature summary. Here, $M$ is the number of model parameters, $r$ is the LoRA rank, $m$ is the weight matrix dimension, $L$ is the number of layers in the LLM, and $K$ is the number of rotating parameter chunks in \textsc{ChunkFT} or the number of layer-wise partitions in LOMO, BAdam, and HiFT. \textit{Dense gradient} denotes whether dense gradients are materialized for the trainable scope. \textit{Optimizer-state continuity} denotes whether historical optimizer states are reused across updates.}
\label{tab:algorithm_summary}
\resizebox{\linewidth}{!}{
\begin{tabular}{lccccccccc}
\toprule
\textbf{Method} 
& \textbf{Memory} 
& \makecell{\textbf{Full param.}\\\textbf{training}}
& \makecell{\textbf{Momentum}\\\textbf{\& 2nd moment}}
& \makecell{\textbf{Update}\\\textbf{precision}}
& \makecell{\textbf{Grad.}\\\textbf{accum.}}
& \makecell{\textbf{Sparse}\\\textbf{gradient}}
& \makecell{\textbf{Optimizer-state}\\\textbf{continuity}}
& \makecell{\textbf{Partition}\\\textbf{upper bound}}
& \makecell{\textbf{Compute per param.}\\\textbf{per full update}} \\
\midrule
Adam& $18M$ & \cmark & \cmark & Float32 & \cmark & \xmark& \cmark& -- & $1\times$ \\

LOMO& $2M + \frac{2M}{K}$ & \cmark & \xmark & Float16 & \xmark & \xmark& \xmark& -- & $1\times$ \\

LoRA& $2M + \frac{36rM}{m}$ & \xmark & \cmark & Float32 & \cmark & \xmark& \cmark& -- & $O(\frac{r}{m})$ \\

BAdam & $2M + \frac{16M}{K}$ & \cmark & \cmark & Float32 & \cmark & \xmark& \xmark& $K \leq L$ & $\frac{K+1}{2}\times$ \\

HiFT & $2M + \frac{16M}{K}$ & \cmark & \cmark & Float32 & \cmark & \xmark& \cmark& $K \leq L$ & $\frac{K+1}{2}\times$ \\
APOLLO 
& $6M + \mathcal O(rM/m)$ & \cmark & \xmark & Float32 & \cmark & \xmark& \cmark& -- & $1\times$ \\

\textsc{ChunkFT} & $2M + \frac{16M}{K}$ & \cmark & \cmark & Float32 & \cmark & \cmark& \cmark& $K \gg L$ & $\approx 1\times$ \\
\bottomrule
\end{tabular}
}
\vspace{-2.0em}
\end{table*}

\section{Related work}
\paragraph{Parameter-Efficient Fine-Tuning.} LoRA (Low-Rank Adaptation) reduces memory usage by optimizing low-rank adaptation matrices instead of all pretrained weights~\cite{HuSWALWWC22}. By restricting updates to a low-dimensional subspace, LoRA substantially lowers the number of trainable parameters and preserves optimizer states for the adapted parameters. Several LoRA variants have been proposed to enhance performance and support multi-task learning~\cite{xia2024chain,tian2024hydralora,liu2026high,hu2026efficient}. Although these methods can achieve competitive results, their limited representation space constrains model performance~\cite{liu2025look}.

\paragraph{Memory-efficient Full Fine-Tuning.} MEFT typically optimizes either the optimizer design or the backpropagation process, reducing the memory footprint of gradients, optimizer states, and activations to enable efficient full-parameter fine-tuning~\cite{lv2024full,zhaogalore}. LOMO~\cite{lv2024full} reduces memory usage by fusing gradient computation with parameter updates and avoiding the storage of conventional optimizer states. While this enables low-memory FFT, it sacrifices momentum and second-moment estimates and does not naturally support gradient accumulation, which may affect optimization stability. BAdam~\cite{luo2024badam} and HiFT~\cite{liu2024hift} further explore partition-based training, where only a subset of parameters or layers is updated at each step.
Recent work also studies optimizer-state compression~\cite{zhang2024adam,zhu2025apollo,dettmers8}. APOLLO~\cite{zhu2025apollo} reduces optimizer memory by using low-rank or projected optimizer statistics, thereby improving memory efficiency while retaining full-parameter training capability.  

As shown in Table~\ref{tab:algorithm_summary}, \textsc{ChunkFT} partitions parameters into fine-grained rotating chunks and updates only a sparse subset of parameters at each step. Unlike LOMO, \textsc{ChunkFT} retains momentum and second-moment states and supports gradient accumulation. Unlike BAdam and HiFT, its partition granularity is not restricted to layer-wise decomposition, allowing the number of chunks (K) to be much larger than the number of layers (L). As a result, \textsc{ChunkFT} combines full-parameter trainability, optimizer-state continuity, sparse gradients, and controllable memory usage, while keeping a computational cost per parameter—averaged over a full parameter update approximately comparable to standard FFT. These properties make \textsc{ChunkFT} an alternative to existing memory-efficient fine-tuning techniques.

\begin{algorithm}[t]
\small
\caption{ChunkFT: Rotating Chunk Full-Parameter Optimization}
\label{alg:mrcopt}
\begin{algorithmic}[1]
\REQUIRE Model parameters $\theta$, number of chunks $K$, loss $\mathcal L$, learning rate $\eta$, AdamW hyperparameters $(\beta_1,\beta_2,\epsilon,\lambda)$, interval counter: $T$
\STATE Estimate byte-level training cost $B(\theta_l)$ for each parameter tensor $\theta_l$
\STATE Partition trainable parameters into byte-balanced chunks $\{\mathcal C_k\}_{k=0}^{K-1}$ based on $B(\theta_l)$
\FOR{$k=0$ to $K-1$}
    \STATE Initialize CPU-resident fp32 master weights $\theta_{\mathrm{CPU}}^{(k)}$ \qquad \qquad \quad // \textit{for mixed precision}
    \STATE Initialize CPU-resident AdamW states $m_{\mathrm{CPU}}^{(k)}, v_{\mathrm{CPU}}^{(k)}$ \qquad \qquad // \textit{supports other optimizers}
    \STATE Set local update counter $n^{(k)} \leftarrow 0$
\ENDFOR
\STATE Add chunk-aware backpropagation for modules such as Embedding, Linear, and LayerNorm
\STATE Initialize active index $\mathcal{I}\leftarrow 0$
\FOR{training step $t=0,1,2,\ldots$}
    \IF{$t \bmod T = 0$}
        \STATE Select active chunk $k \leftarrow \mathcal I \bmod K$
        \STATE Asynchronously load $\theta_{\mathrm{CPU}}^{(k)}, m_{\mathrm{CPU}}^{(k)}, v_{\mathrm{CPU}}^{(k)}$ to GPU
        \STATE Enable gradient accumulation only for parameter slices $\theta^{(k)}$ in $\mathcal C_k$
    \ENDIF
    \STATE Compute prediction $\hat y \leftarrow f(x;\theta)$
    \STATE Compute loss $\ell \leftarrow \mathcal L(\hat y, y)$
    \STATE Run chunk-aware backward to obtain $g^{(k)} = \nabla_{\theta^{(k)}} \ell$ instead of dense $\theta_{\mathrm{grad}}$
    \STATE $n^{(k)} \leftarrow n^{(k)} + 1$;\qquad$m^{(k)} \leftarrow \beta_1 m^{(k)} + (1-\beta_1)g^{(k)}$
    \STATE $v^{(k)} \leftarrow \beta_2 v^{(k)} + (1-\beta_2)(g^{(k)})^2$;\qquad$\hat m^{(k)} \leftarrow m^{(k)} / (1-\beta_1^{n^{(k)}})$
    \STATE $\hat v^{(k)} \leftarrow v^{(k)} / (1-\beta_2^{n^{(k)}})$;\qquad$\theta^{(k)} \leftarrow \theta^{(k)} - \eta \left(\hat m^{(k)} / (\sqrt{\hat v^{(k)}}+\epsilon) + \lambda \theta^{(k)}\right)$
    \STATE Clear temporary chunk gradients
    \IF{$t \bmod T = T-1$}
        \STATE Asynchronously offload updated $\theta^{(k)}, m^{(k)}, v^{(k)}$ to CPU
        \STATE Release GPU-resident states of chunk $k$ after offload completes
        \STATE $I \leftarrow I + 1$
    \ENDIF
\ENDFOR
\end{algorithmic}
\end{algorithm}
\vspace{-1em}

\section{Method}
\label{sub:method}
We present \textsc{ChunkFT}, a memory-centric optimization framework for large-scale transformer adaptation. For each parameter tensor, we estimate its size as numel $\times$ element\_size and accumulate parameters in model order until a predefined byte budget is reached. Large matrices that exceed the budget are split along the row dimension, allowing one tensor to span multiple chunks. During training, only the active chunk is updated while the others remain frozen.
This yields a training regime in which memory is organized around an active working set rather than the full trainable parameter space.

\subsection{Algorithm Description}
Algorithm~\ref{alg:mrcopt} summarizes the training procedure of \textsc{ChunkFT}. 
Given trainable parameters $\theta$, \textsc{ChunkFT} first estimates the byte-level training cost $B(\theta_l)$ for each parameter tensor $\theta_l$, including parameter storage, gradient buffer, fp32 master copy, and optimizer states. 
The trainable parameters are then partitioned into $K$ byte-balanced chunks $\{\mathcal C_k\}_{k=0}^{K-1}$, so that each chunk induces a comparable training-time memory footprint. For each chunk $\mathcal C_k$, \textsc{ChunkFT} initializes CPU-resident fp32 master weights for mixed precision training and optimizer states. For AdamW, these states are the first- and second-moment estimates $m_{\mathrm{CPU}}^{(k)}$ and $v_{\mathrm{CPU}}^{(k)}$. 
A local update counter $n^{(k)}$ is maintained for each chunk for bias correction, since chunks are updated at different global steps.

During training, \textsc{ChunkFT} follows a rotating schedule. Given a chunk update interval $T$, the same chunk is optimized for $T$ consecutive steps before switching to the next one in a round-robin order. 
When chunk $\mathcal C_k$ becomes active, its fp32 master weights and optimizer states are asynchronously loaded from CPU to GPU, and gradient accumulation is enabled only for the parameter slices in $\mathcal C_k$. The forward pass still uses the full model, $\hat y = f(x;\theta)$,
so the training objective remains identical to standard full-parameter fine-tuning. 
However, the backward pass is chunk-aware: instead of materializing dense gradients for all trainable parameters, it computes only
\begin{equation}
g^{(k)} = \nabla_{\theta^{(k)}} \mathcal L(\hat y,y),
\end{equation}
where $\theta^{(k)}$ denotes the active parameter slices in $\mathcal C_k$. 
Inactive chunks do not allocate gradient buffers. 
This is implemented by chunk-aware backward operators for memory-dominant modules such as Embedding, Linear, and LayerNorm. The optimizer update is then applied only to the active chunk. 

All inactive chunks remain unchanged. After the active chunk has been updated for $T$ steps, its updated master weights and optimizer states are asynchronously offloaded back to CPU, GPU buffers are released, and the active index advances to the next chunk. Thus, \textsc{ChunkFT} performs full-parameter optimization over time while requiring gradient and optimizer-state residency only for the active chunk at each step. 
With byte-balanced chunks, the peak optimizer-related GPU memory is approximately
\begin{equation}
\max_k B(\mathcal C_k)
\approx
\frac{1}{K}
\sum_l B(\theta_l),
\end{equation}
up to temporary transfer and buffering overhead.

It is important to note that \textsc{ChunkFT} differs from existing BAdam methods \cite{luo2024badam}.
BAdam still follows a layer-wise optimization paradigm, and inactive blocks are masked by gradient masking after gradient computation. 
Similar to layer-wise fine-tuning methods like HiFT \cite{liu2024hift},
its
main mechanism is to skip the optimizer update for inactive parameters rather than to prevent their gradients from being generated.

\paragraph{Convergence result.}
We provide a convergence analysis for \textsc{ChunkFT} in the deterministic case, aiming to establish that rotating chunk optimization yields a convergent block-coordinate descent scheme. 
The formal theorem and proof are provided in Appendix~\ref{app:chunkft_convergence}; here we state the 
key takeaways.



\begin{theorem}
\label{thm:chunkft_informal}
Let the trainable parameters be partitioned into $K$ chunks. \textsc{ChunkFT} cyclically updates one active chunk at a time, performing $T$ inner updates for each active chunk. Let $R$ denote the number of completed chunk rotations, where one rotation means that all $K$ chunks have been updated once. For a sufficiently small learning rate, the average active-chunk gradient satisfies
\begin{equation}
\frac{1}{RTK}
\sum_{r=0}^{R-1}
\sum_{i=1}^{K}
\sum_{t=0}^{T-1}
\left\|
\nabla_i \mathcal L(\theta_i^{r,t})
\right\|^2
=
\mathcal O\left(\frac{1}{RTK}\right).
\end{equation}
Consequently, over completed chunk rotations, \textsc{ChunkFT} converges to a first-order stationary point. For fixed $K$ and $T$, it reaches a $\delta$-approximate stationary point within $\mathcal O(\delta^{-2})$ active-chunk updates.
\end{theorem}

This result shows that \textsc{ChunkFT} has the same convergence behavior as block coordinate descent.

\subsection{BP Time Analysis}
\label{subsec:bp_time_analysis}

We analyze the backward computation required to generate parameter gradients. 
We use one \emph{full-parameter cycle} as the comparison unit, defined as the minimum number of steps required for all trainable parameters to be selected for update once. Let $C$ denote the cost of generating dense gradients for all trainable parameters once. For standard full-parameter backpropagation, one full-parameter cycle costs $C_{\mathrm{full}} = C$. \textsc{ChunkFT} partitions trainable parameters into \(K\) disjoint chunks: $\{\mathcal C_1,\mathcal C_2,\ldots,\mathcal C_K\}$.
Let \(C_k\) denote the cost of generating gradients for chunk \(\mathcal C_k\). 
Since the chunks form a disjoint partition of the trainable parameter space, $\sum_{k=1}^{K} C_k = C$.
At each step, \textsc{ChunkFT} activates one chunk and uses chunk-aware backward operators to generate gradients only for the active parameter slices. 
Therefore, the gradient-generation cost of updating chunk \(\mathcal C_k\) is $C_{\textsc{ChunkFT}}(k)=C_k$.
Over one full rotation, all chunks are activated once. 
Thus, the total gradient-generation cost is
\begin{equation}
C_{\textsc{ChunkFT}}
=
\sum_{k=1}^{K} C_k
=
C.
\end{equation}
After normalization by dense full-parameter gradient generation, we obtain $\frac{C_{\textsc{ChunkFT}}}{C_{\mathrm{full}}}=1$. This shows that \textsc{ChunkFT} requires only one dense gradient-generation pass per full-parameter cycle. 
The key difference is that \textsc{ChunkFT} distributes this computation across \(K\) steps and never materializes dense gradients at an individual step. For a comprehensive analysis of other methods, see Appendix~\ref{app:bp_time_comparison}.

\subsection{Memory Consumption Analysis}
We analyze the memory usage of \textsc{ChunkFT}. Consider a model with \(M\) billion trainable parameters, and measure memory in GB. With mixed-precision AdamW, FP16 model parameters require \(2M\) memory, while the FP32 master weights, gradients, first-moment states, and second-moment states each require \(4M\). 
Thus, the total optimizer-related memory is approximately
\begin{equation}
2M + 4M + 4M + 4M + 4M = 18M .
\end{equation}
In contrast, \textsc{ChunkFT} keeps the full up-to-date FP16 model on GPU, costing \(2M\), but materializes FP32 gradients, master weights, and AdamW states only for the active chunk. 
If the parameters are divided into \(K\) equal-sized chunks, the active chunk contains \(M/K\) parameters, so these active-chunk states cost
\begin{equation}
4\frac{M}{K}
+4\frac{M}{K}
+4\frac{M}{K}
+4\frac{M}{K}
=
\frac{16M}{K}.
\end{equation}
Therefore, the total memory required by \textsc{ChunkFT} is $2M+\frac{16M}{K}$.

Unlike layer-wise partitioning, \textsc{ChunkFT} partitions parameters by a byte-level memory budget rather than by layer identity. 
As a result, \(K\) is not limited by the number of transformer layers ($K_{\mathrm{ChunkFT}} \gg K_{\mathrm{BAdam}}$), and the per-chunk GPU memory pressure is more balanced.

\begin{table*}[ht]
\captionsetup{skip=2pt}
\caption{Actual memory costs of fine-tuning Llama 2-7B with gradient checkpointing on the BoolQ~\cite{clark2019boolq} dataset. The maximum input sequence length is 1024. The reported results do not use DeepSpeed’s memory optimization techniques. MP indicates mixed precision.}
\label{tab:memory}
\begin{adjustbox}{max width=0.85\textwidth, center}
\includegraphics[width=\textwidth]{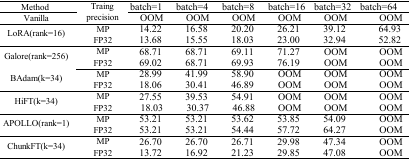}
\end{adjustbox}
\vspace{-2.0em}
\end{table*}

\begin{table*}[ht]
\captionsetup{skip=2pt}
\caption{Time spent fine-tuning Llama 2-7B with gradient checkpointing on BoolQ dataset (with 1000 training examples and 500 validation examples). All results are obtained from experiments conducted on a single H800-80GB.}
\label{tab:speed}
\begin{adjustbox}{max width=0.80\textwidth, center}
\includegraphics[width=\textwidth]{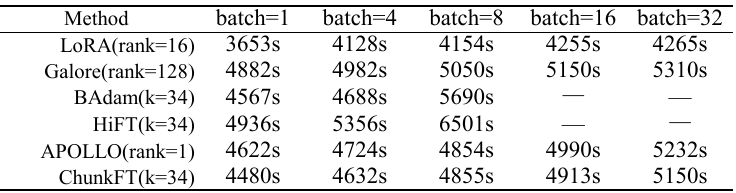}
\end{adjustbox}
\vspace{-2.0em}
\end{table*}

\begin{table*}[ht]
\captionsetup{skip=2pt}
\caption{SuperGLUE~\cite{wang2019superglue} benchmark scores of RoBERTa-large (standard deviation).}
\label{tab:roblarge}
\begin{adjustbox}{max width=0.80\textwidth, center}
\includegraphics[width=\textwidth]{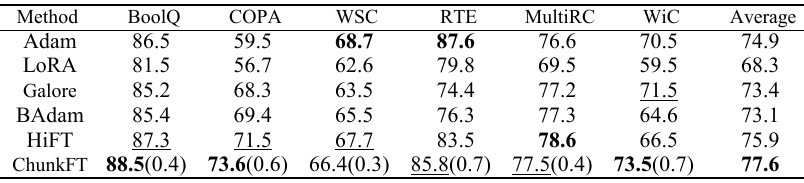}
\end{adjustbox}
\vspace{-2.0em}
\end{table*}
\section{Experiment Results}

In this section, we evaluate the proposed \textsc{ChunkFT} for fine-tuning LLMs. The baselines include
Adam~\cite{kingma2014adam}, LOMO~\cite{lv2024full}, LoRA~\cite{hu2022lora}, GaLore~\cite{zhaogalore}, HiFT~\cite{liu2024hift}, BAdam~\cite{luo2024badam}, and the recent method APOLLO~\cite{zhu2025apollo}. All \textsc{ChunkFT} experiments
for RoBERTa-large~\cite{liu2019roberta} are conducted on a single RTX 4090-24GB GPU, while the others are conducted on H800-80GB GPUs. Our implementation is based on the Transformers library. The detailed experiment setup can be found in Appendix~\ref{app:implement}.

\subsection{Memory Consumption and Wall-clock Running Time}
\label{sub:memory-time}
\paragraph{Memory consumption.}
We report the actual memory consumption of different fine-tuning methods in Table~\ref{tab:memory}. LoRA consumes the least GPU memory due to its non-full-parameter fine-tuning setting. Surprisingly, \textsc{ChunkFT} achieves memory consumption comparable to LoRA, while requiring the least memory among all full-parameter fine-tuning methods.  Compared with GaLore, BAdam, HiFT, and APOLLO, \textsc{ChunkFT} supports a maximum batch size of 8 on a 24GB GPU for benchmarks with a sequence length of 1024, surpassing all baseline MEFT methods. Under FP32 precision, \textsc{ChunkFT} consumes only 13.72GB at batch size 1, which is comparable to LoRA, while it remains feasible at batch size 32 where BAdam and HiFT already run out of memory. These results demonstrate that \textsc{ChunkFT} effectively controls the memory costs, making it practical for fine-tuning large language models under memory-constrained settings.
\begin{wraptable}{r}{0.45\columnwidth}
  \vspace{-0.5em}
  \centering
  \captionsetup{skip=2pt}
  \caption{Comparison of memory fluctuations across different fine-tuning methods on BoolQ dataset (batch size is 8).}
    \label{tab:jitter}
    \includegraphics[width=\linewidth]{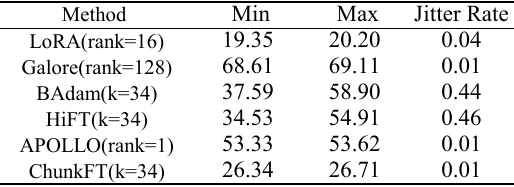}
\vspace{-2.0em}
\end{wraptable}
Mixed precision does not always lead to lower memory consumption in our experiments. At small batch sizes, mixed precision can even consume more memory than FP32. This is mainly because automatic mixed-precision training maintains additional FP32 master weights, scaling-related buffers, and casted intermediate tensors for numerical stability, while the activation memory is relatively small when the batch size is low. As a result, the extra overhead introduced by mixed precision may outweigh the memory saved by using lower-precision activations. Nevertheless, as the batch size increases, activation memory becomes more dominant, and mixed precision becomes more effective in reducing the overall memory footprint.

\paragraph{Wall-clock running time comparison.}
As shown in Table~\ref{tab:speed}, we conduct experiments on fine-tuning the Llama 2-7B model for three epochs with each method and report the average wall-clock time per epoch. Overall, \textsc{ChunkFT} demonstrates strong efficiency and scalability across different batch sizes. Although LoRA achieves the shortest training time, this is expected since LoRA only updates a small set of low-rank adapter parameters. Compared with BAdam and HiFT, \textsc{ChunkFT} shows clear advantages. Under comparable settings, \textsc{ChunkFT} is consistently faster than BAdam and HiFT. For instance, at batch size 8, \textsc{ChunkFT} takes 4855 seconds, compared with 5690 seconds for BAdam and 6501 seconds for HiFT.

ChunkFT is also competitive with GaLore and APOLLO. It is faster than GaLore across all reported batch sizes. At batch size 32, \textsc{ChunkFT} takes 5150 seconds, while GaLore requires 5310 seconds. Compared with APOLLO, \textsc{ChunkFT} has similar runtime and is slightly faster at larger batch sizes. For example, \textsc{ChunkFT} takes 4913 seconds and 5150 seconds at batch sizes 16 and 32, respectively, compared with 4990 seconds and 5232 seconds for APOLLO. These results indicate that the chunk-wise mechanism of \textsc{ChunkFT} does not introduce substantial additional overhead and can even improve efficiency. This is mainly due to the efficient design of ChunkFT. Unlike HiFT and BAdam, where gradients propagate through the entire computation graph, including inactive layers, \textsc{ChunkFT} restricts gradient flow to the active sub-tensors. This substantially reduces unnecessary computation and avoids wasted gradient calculations.
\begin{table*}[ht]
\captionsetup{skip=2pt}
\caption{SuperGLUE~\cite{wang2019superglue} benchmark scores of Llama 2-7B using different optimization methods.}
\label{tab:llama2-7B}
\begin{adjustbox}{max width=0.85\textwidth, center}
\includegraphics[width=\textwidth]{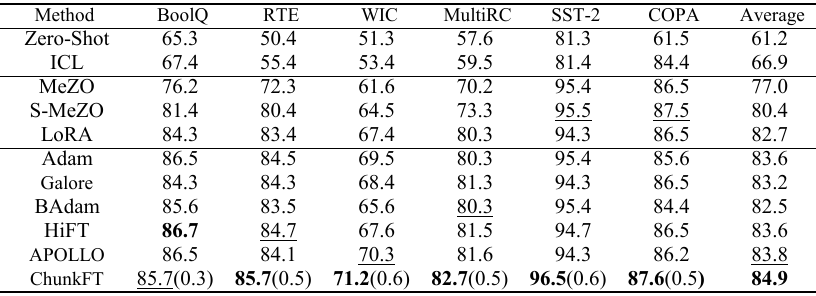}
\end{adjustbox}
\vspace{-2.0em}
\end{table*}
\paragraph{Memory jitter comparison.}
Table~\ref{tab:jitter} compares the memory fluctuation of different fine-tuning methods. Overall, \textsc{ChunkFT} exhibits highly stable memory usage, with memory varying only from 26.34GB to 26.71GB and a jitter rate of 0.01. This is comparable to LoRA, GaLore, and APOLLO, which also achieve a jitter rate of 0.01, while requiring substantially less memory than both methods. In particular, \textsc{ChunkFT} reduces the maximum memory consumption by 61.35\% compared with GaLore and by 50.18\% compared with APOLLO.

In contrast, BAdam and HiFT show much larger memory fluctuations. BAdam has a jitter rate of 0.44, with memory increasing from 37.59GB to 58.90GB, while HiFT has a jitter rate of 0.46, increasing from 34.53GB to 54.91GB. In LLMs, parameter sizes can vary substantially across layers/blocks, which is a major cause of memory jitter in HiFT and BAdam. \textsc{ChunkFT} maintains a much narrower memory range, suggesting that its stream mechanism effectively mitigates memory thrashing caused by parameter-size variations across different layers/blocks.
\begin{table*}[ht]
\captionsetup{skip=2pt}
\caption{Math benchmark scores of Llama 3-8B using different optimization methods.}
\label{tab:llama3-8B}
\begin{adjustbox}{max width=0.85\textwidth, center}
\includegraphics[width=\textwidth]{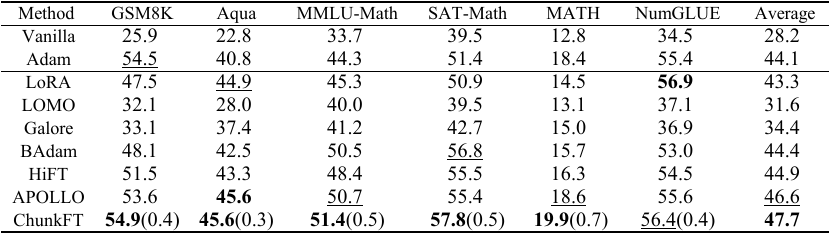}
\end{adjustbox}
\vspace{-2.0em}
\end{table*}

\begin{table*}[ht]
\captionsetup{skip=2pt}
\caption{Math benchmark scores of Llama 3-70B using different optimization methods.}
\label{tab:llama3-70B}
\begin{adjustbox}{max width=0.85\textwidth, center}
\includegraphics[width=\textwidth]{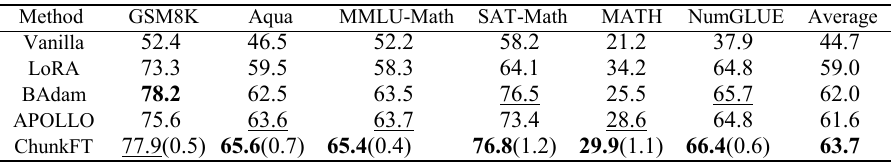}
\end{adjustbox}
\vspace{-2.0em}
\end{table*}
\subsection{Downstream Performance Evaluation.}
\paragraph{Natural language understanding.}
Tables~\ref{tab:roblarge} and~\ref{tab:llama2-7B} report the performance of different optimization methods on natural language understanding benchmarks. 
On RoBERTa-large, \textsc{ChunkFT} obtains the highest average score of 77.6, outperforming Adam, LoRA, GaLore, BAdam, and HiFT. In particular, \textsc{ChunkFT} achieves the best results on BoolQ, COPA, and WiC, with scores of 88.5, 73.6, and 73.5, respectively. Compared with Adam FFT, \textsc{ChunkFT} improves the average score from 74.9 to 77.6, while also yielding notable gains on BoolQ and COPA.
On Llama 2-7B, \textsc{ChunkFT} also achieves the best average score of 84.9. It obtains the highest scores on RTE, WiC, MultiRC, SST-2, and COPA, and remains competitive on BoolQ. Compared with Adam and HiFT, both of which achieve an average score of 83.6, \textsc{ChunkFT} improves the average performance by 1.3 points. It also outperforms APOLLO, GaLore, BAdam, and LoRA, indicating that \textsc{ChunkFT} is not only memory- and time-efficient, but also preserves strong downstream task performance. These results suggest that \textsc{ChunkFT} provides a favorable trade-off between efficiency and effectiveness.
\paragraph{Math benchmarks.}
Tables~\ref{tab:llama3-8B} and~\ref{tab:llama3-70B} report the results on math benchmarks for Llama 3-8B and Llama 3-70B. On Llama 3-8B, \textsc{ChunkFT} attains the highest average score of 47.7, outperforming all baselines, including APOLLO (46.6), HiFT (44.9), BAdam (44.4), and Adam (44.1). Specifically, \textsc{ChunkFT} achieves the best or tied-best performance on 5 out of 6 benchmarks.
The advantage of \textsc{ChunkFT} becomes even clearer on Llama 3-70B. \textsc{ChunkFT} achieves the best overall average of 63.7, surpassing BAdam (62.0), APOLLO (61.6), and LoRA (59.0). It obtains the best results on 5 out of 6 benchmarks. These results suggest that \textsc{ChunkFT} provides a favorable trade-off between optimization efficiency and downstream reasoning performance. Furthermore, the small variances shown in parentheses indicate that \textsc{ChunkFT} is not only effective but also stable across runs. Overall, the results on both Llama 3-8B and Llama 3-70B verify that \textsc{ChunkFT} is a robust and scalable optimization method for mathematical reasoning tasks.
\begin{table*}[ht]
\captionsetup{skip=2pt}
\caption{MT-Bench scores of instruction-tuned Llama 2-7B and Llama 3-8B on Alpaca-GPT4.}
\label{tab:mt-bench}
\begin{adjustbox}{max width=0.9\textwidth, center}
\includegraphics[width=\textwidth]{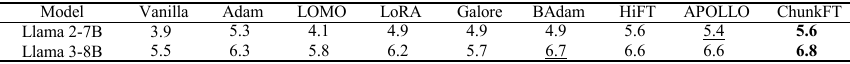}
\end{adjustbox}
\vspace{-2.0em}
\end{table*}
\paragraph{MT-bench results.}

Table~\ref{tab:mt-bench} reports the MT-Bench scores of instruction-tuned Llama 2-7B and Llama 3-8B on Alpaca-GPT4. Overall, \textsc{ChunkFT} achieves the best performance on both models, showing that the proposed method is effective for instruction-following generation. For Llama 2-7B, \textsc{ChunkFT} obtains an MT-Bench score of 5.6, matching HiFT and outperforming Adam, LoRA, GaLore, BAdam, APOLLO, LOMO, and the vanilla model. Compared with Adam FFT, \textsc{ChunkFT} improves the score from 5.3 to 5.6. For Llama 3-8B, \textsc{ChunkFT} achieves the highest score of 6.8, outperforming all baselines. In particular, it improves over Adam by 0.5 points, over LoRA by 0.6 points, and over APOLLO by 0.2 points. Although BAdam and HiFT also achieve strong performance, \textsc{ChunkFT} still provides the best overall result. These results demonstrate that \textsc{ChunkFT} achieves top MT-Bench scores, confirming its effectiveness as a practical optimization method for instruction tuning.
\begin{figure*}[t]
    \centering
    \captionsetup{skip=2pt}
    \includegraphics[width=0.8\textwidth]{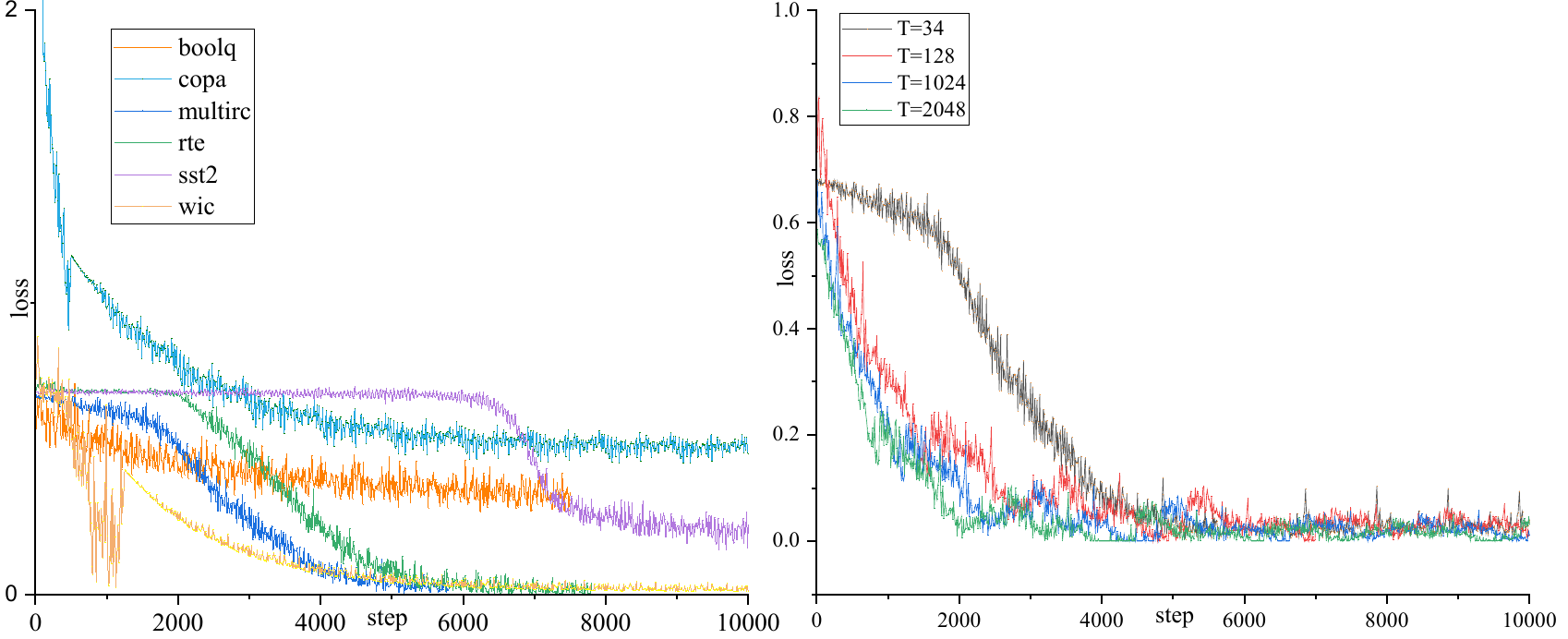}
    \caption{Loss convergence behavior of \textsc{ChunkFT} (Llama 2-7B). \textbf{Left:} training loss curves of \textsc{ChunkFT} on six natural language understanding tasks. \textbf{Right:} loss curves of training a single block with different update intervals $T$ on BoolQ.}
    \label{fig:conve_loss}
\vspace{-1.0em}
\end{figure*}
\begin{figure*}[t]
    \centering
    \captionsetup{skip=2pt}
    \includegraphics[width=0.8\textwidth]{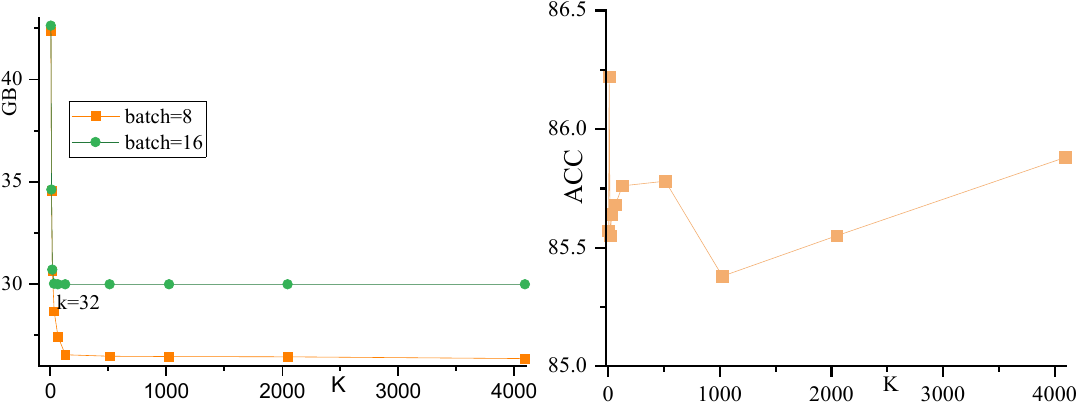}
    \caption{Effect of the chunk number $K$ on memory and performance. \textbf{Left:} peak GPU memory under different chunk numbers $K$. \textbf{Right:} BoolQ accuracy under different chunk numbers $K$.}
    \vspace{-1.0em}
    \label{fig:block-abla}
\end{figure*}
\subsection{Ablation Study}
\paragraph{Loss convergence.}
Figure~\ref{fig:conve_loss} (left) shows the training loss curves of \textsc{ChunkFT} on six natural language understanding tasks. Overall, \textsc{ChunkFT} exhibits stable convergence behavior across all tasks, with the loss decreasing steadily throughout training. Specifically, SST-2, WiC, and COPA converge relatively quickly to near-zero loss, while BoolQ and RTE decrease more gradually and plateau at higher loss values. MultiRC also shows a slower decay pattern, reflecting its relatively higher complexity. These results indicate that the chunk-wise optimization strategy of \textsc{ChunkFT} maintains stable and effective training dynamics across a diverse set of tasks.

\paragraph{Update intervals.}
Figure~\ref{fig:conve_loss} (right) studies the effect of the update interval $T$ when optimizing a single block. The results show that different choices of $T$ do not substantially affect the overall convergence trend: all settings eventually converge to a low loss region and exhibit stable optimization behavior. This indicates that \textsc{ChunkFT} is not sensitive to the exact update interval in terms of final convergence. However, the update interval can influence the convergence speed. When $T$ is small, such as $T{=}34$, blocks are switched more frequently, which may interrupt local optimization and slow down the early-stage loss decrease. In contrast, larger intervals allow each active block to be optimized for more consecutive steps, leading to faster initial convergence. Therefore, while the update interval has limited impact on whether \textsc{ChunkFT} converges, overly frequent block switching may reduce optimization efficiency.

\paragraph{Impact of chunk number on memory.}
Figure~\ref{fig:block-abla} (left) shows the effect of the chunk number $K$ on GPU memory usage for Llama 2-7B on BoolQ under batch sizes 8 and 16. 
Overall, increasing $K$ significantly reduces memory consumption. Moreover, after $K$ reaches a moderate value, the memory footprint becomes nearly flat, suggesting that further increasing the number of chunks brings diminishing returns. The reason is that, as the number of chunks increases, fewer parameters are active at each step, which reduces the amount of optimizer state and intermediate memory that needs to be materialized simultaneously. Such small variations are negligible in practice.
\paragraph{Impact of chunk number on performance.}

Figure~\ref{fig:block-abla} (right) presents the corresponding BoolQ accuracy under different chunk numbers $K$. We observe that the performance remains highly stable across a wide range of chunk settings. Specifically, all results fall within a narrow range of less than 1\% point, which is within normal experimental fluctuation. Although there are small variations across different $K$ values, no clear degradation trend is observed as the model is divided into more chunks. This suggests that the chunk-wise training strategy does not materially harm optimization quality or downstream generalization. Combined with the memory results, these findings indicate that \textsc{ChunkFT} achieves a favorable trade-off: it substantially reduces memory usage while maintaining essentially unchanged task performance.

\section{Conclusion}

In this paper, we propose ChunkFT, an efficient and scalable fine-tuning method for large language models. \textsc{ChunkFT} partitions model parameters into byte-stream chunks and selectively activates only a subset of parameters during each update, thereby reducing unnecessary gradient computation and memory consumption. Unlike methods whose gradient flow spans the entire computation graph, \textsc{ChunkFT} localizes optimization to active sub-tensors. Its byte-stream allocation further balances chunk sizes and mitigates memory jitter caused by layer-wise parameter variations. In addition, \textsc{ChunkFT} supports gradient updates on arbitrary sub-tensors, decoupling the update granularity from the model architecture and enabling flexible, stable, and resource-efficient fine-tuning.


\bibliography{custom}

\clearpage

\appendix

\section{Benchmarks}
\label{app:benchmarks}
To comprehensively evaluate the capabilities of different methods, we conduct experiments on a diverse set of benchmarks covering \textbf{mathematical reasoning}, \textbf{natural language understanding}, and \textbf{MT-Bench}~\cite{zheng2023judging}.
Specifically, the mathematical reasoning benchmarks assess models' ability to solve arithmetic, algebraic, numerical, and competition-level problems through multi-step reasoning. The natural language understanding benchmarks evaluate
core linguistic abilities, including reading comprehension, textual entailment, commonsense reasoning, coreference resolution, word sense disambiguation, and sentiment analysis. In addition, we include MT-Bench to measure open-ended
dialogue quality and instruction-following performance in multi-turn conversational scenarios. Together, these benchmarks provide a comprehensive evaluation suite that reflects both task-specific reasoning ability and general
assistant-oriented language understanding and generation capability.

\paragraph{Mathematical Reasoning Datasets.}
We evaluate mathematical reasoning ability on six representative benchmarks:
GSM8K~\cite{cobbe2021training}, AQuA~\cite{ling2017program}, MMLU-Math~\cite{hendrycks2020measuring}, SAT-Math~\cite{zhong2024agieval}, MATH~\cite{hendrycks2021measuring}, and NumGLUE~\cite{mishra2022numglue}. These datasets cover a
wide range of mathematical reasoning scenarios, from elementary arithmetic word
problems to advanced competition-level problem solving and numerical reasoning
in natural language.

\begin{itemize}[leftmargin=*]
    \item \textbf{GSM8K.}
    GSM8K~\cite{cobbe2021training} is a grade-school mathematics benchmark consisting of natural
    language word problems. Each problem typically requires multiple steps of
    arithmetic reasoning, such as addition, subtraction, multiplication,
    division, and proportional reasoning. Since the questions are presented in
    everyday scenarios, GSM8K evaluates not only the model's calculation ability
    but also its capacity to understand problem contexts, identify relevant
    quantities, and generate coherent intermediate reasoning steps.

    \item \textbf{AQuA.}
    AQuA~\cite{ling2017program} is an algebraic question answering dataset composed mainly of
    multiple-choice math problems. The questions involve topics such as algebra,
    equations, ratios, probability, and numerical computation. Solving AQuA
    problems often requires translating natural language descriptions into
    mathematical expressions, applying symbolic manipulation or equation
    solving, and selecting the correct answer from candidate options. It is
    commonly used to evaluate mathematical reasoning under a standardized
    multiple-choice setting.

    \item \textbf{MMLU-Math.}
    MMLU-Math~\cite{hendrycks2020measuring} refers to the mathematics-related subset of the MMLU benchmark.
    It covers a broad spectrum of mathematical knowledge, including elementary
    mathematics, high-school mathematics, abstract algebra, and formal logic.
    Compared with datasets focused mainly on arithmetic computation,
    MMLU-Math emphasizes mathematical concept understanding and domain
    knowledge across different levels of difficulty. It is therefore suitable
    for assessing a model's general mathematical knowledge and reasoning
    ability.

    \item \textbf{SAT-Math.}
    SAT-Math~\cite{zhong2024agieval} contains math questions derived from standardized SAT-style
    examinations. The problems cover common exam topics such as algebra,
    geometry, functions, statistics, probability, and data analysis. These
    questions typically require concise reasoning, accurate application of
    mathematical rules, and familiarity with standardized test formats.
    SAT-Math evaluates whether a model can solve structured exam-style
    problems that resemble real educational assessment scenarios.

    \item \textbf{MATH.}
    MATH~\cite{hendrycks2021measuring} is a challenging mathematical reasoning benchmark consisting of
    competition-level problems. It spans diverse mathematical areas, including
    algebra, number theory, geometry, combinatorics, probability, precalculus,
    and calculus. Unlike simpler arithmetic datasets, MATH often requires
    advanced symbolic reasoning, multi-step derivations, theorem application,
    and sometimes proof-like problem solving. As a result, it serves as a
    strong benchmark for evaluating the upper-level mathematical reasoning
    capability of large language models.

    \item \textbf{NumGLUE.}
    NumGLUE~\cite{mishra2022numglue} is a benchmark designed to evaluate numerical reasoning in natural
    language understanding tasks. It focuses on the model's ability to
    recognize and compare quantities, perform arithmetic operations, infer
    numerical relations, and integrate numerical information with textual
    semantics. NumGLUE is particularly useful for testing whether models can
    handle numbers accurately in realistic language contexts, rather than only
    solving explicitly formatted math problems.
\end{itemize}

\paragraph{Natural Language Understanding Datasets.}
We evaluate natural language understanding ability on several representative
benchmarks, including BoolQ~\cite{clark2019boolq}, COPA~\cite{roemmele2011choice}, WSC~\cite{levesque2012winograd}, RTE~\cite{dagan2005pascal,bentivogli2009fifth}, MultiRC~\cite{khashabi2018looking}, WiC~\cite{pilehvar2019wic}, and SST-2~\cite{socher2013recursive}. These
datasets cover a broad range of language understanding scenarios, such as
question answering, commonsense reasoning, coreference resolution, textual
entailment, reading comprehension, word sense disambiguation, and sentiment
classification.

\begin{itemize}[leftmargin=*]
    \item \textbf{BoolQ.}
    BoolQ~\cite{clark2019boolq} is a binary question answering dataset in which each example consists
    of a passage, a naturally occurring question, and a boolean answer. The
    questions are typically derived from real user information-seeking queries,
    and solving them requires models to understand the given passage, locate
    relevant evidence, and determine whether the answer is true or false. BoolQ
    is commonly used to evaluate reading comprehension and factual reasoning
    over short passages.

    \item \textbf{COPA.}
    COPA~\cite{roemmele2011choice}, the Choice of Plausible Alternatives dataset, is designed to evaluate
    commonsense causal reasoning. Each instance provides a premise and two
    candidate alternatives, where the model must select the more plausible
    cause or effect. Since the correct answer often depends on everyday
    commonsense knowledge rather than explicit lexical overlap, COPA is useful
    for assessing whether models can reason about causal relations and real-world
    events.

    \item \textbf{WSC.}
    WSC~\cite{levesque2012winograd}, the Winograd Schema Challenge, focuses on coreference resolution under
    challenging linguistic conditions. Each example contains a sentence with an
    ambiguous pronoun and two possible antecedents, and the model must identify
    the correct referent. WSC examples are designed so that simple syntactic or
    statistical cues are insufficient, requiring commonsense reasoning and
    careful semantic understanding.

    \item \textbf{RTE.}
    RTE~\cite{wang2019superglue}, Recognizing Textual Entailment, is a binary classification task that
    determines whether a hypothesis can be inferred from a given premise. The
    dataset evaluates a model's ability to understand semantic relations between
    sentence pairs, including entailment and non-entailment. It is widely used
    to assess natural language inference ability and logical consistency in
    language understanding.

    \item \textbf{MultiRC.}
    MultiRC~\cite{khashabi2018looking} is a multi-sentence reading comprehension dataset. Each instance
    contains a paragraph, a question, and multiple candidate answers, where more
    than one answer may be correct. Solving MultiRC often requires integrating
    evidence from multiple sentences, performing compositional reasoning, and
    distinguishing partially correct answers from fully supported ones. This
    makes it suitable for evaluating complex reading comprehension beyond
    single-sentence evidence matching.

    \item \textbf{WiC.}
    WiC~\cite{pilehvar2019wic}, Word-in-Context, is a word sense disambiguation benchmark. Each example
    presents two sentences containing the same target word, and the model must
    determine whether the word is used with the same meaning in both contexts.
    WiC evaluates fine-grained lexical semantic understanding, requiring models
    to distinguish subtle sense differences based on contextual usage.

    \item \textbf{SST-2.}
    SST-2~\cite{socher2013recursive}, the Stanford Sentiment Treebank binary classification task, evaluates
    sentiment analysis ability. Each example consists of a sentence from movie
    reviews, and the model must classify it as expressing either positive or
    negative sentiment. SST-2 tests whether models can recognize affective
    polarity, handle sentiment-bearing expressions, and perform robust sentence-
    level classification.
\end{itemize}

\paragraph{MT-Bench Dataset.}
MT-Bench~\cite{zheng2023judging} is a multi-turn instruction-following benchmark designed to evaluate
the conversational ability of large language models in open-ended dialogue
scenarios. Each example typically consists of two turns of user instructions,
where the second turn often depends on the context established in the first
turn. The benchmark covers diverse task categories, including writing, role
play, reasoning, mathematics, coding, extraction, humanities, and STEM-related
questions, thereby testing models' abilities in instruction following,
context retention, logical reasoning, factual understanding, and response
generation. Since MT-Bench contains open-ended questions with multiple possible
valid answers, it is commonly evaluated using an LLM-as-a-judge protocol, where
a strong evaluator model scores responses according to helpfulness, relevance,
accuracy, coherence, and instruction adherence. Overall, MT-Bench provides a
practical evaluation setting for measuring whether large language models can
serve as capable conversational assistants across diverse multi-turn
interaction scenarios.

\section{Backbone Models}

\label{app:backbone}
We conduct experiments on four representative backbone models, including Llama 3-8B~\cite{grattafiori2024llama}, Llama 3-70B~\cite{grattafiori2024llama}, Llama 2-7B~\cite{touvron2023llama}, and RoBERTa-large~\cite{liu2019roberta}. These models cover both decoder-only generative language models and encoder-only discriminative
language models, enabling evaluation across different model architectures, parameter scales, and pretraining paradigms.

\begin{itemize}[leftmargin=*]
    \item \textbf{Llama 3-8B.}
    Llama 3-8B~\cite{grattafiori2024llama} is an 8-billion-parameter decoder-only large language model from
    the Llama 3 family. It is designed for general-purpose language
    understanding and generation, and provides a strong balance between
    performance and computational efficiency. Due to its moderate parameter
    size, Llama 3-8B is suitable for evaluating reasoning, instruction following,
    and generation ability under relatively practical resource constraints.

    \item \textbf{Llama 3-70B.}
    Llama 3-70B~\cite{grattafiori2024llama} is a larger model from the Llama 3 family, containing
    approximately 70 billion parameters. Compared with Llama 3-8B, it has
    stronger model capacity and typically demonstrates better performance on
    complex tasks such as mathematical reasoning, commonsense reasoning, and
    long-form generation. We include Llama 3-70B to examine how the proposed
    method performs on a large-scale high-capacity language model.

    \item \textbf{Llama 2-7B.}
    Llama 2-7B~\cite{touvron2023llama} is a 7-billion-parameter decoder-only language model from the
    Llama 2 family. As an earlier-generation open-weight large language model,
    it serves as a widely used baseline for evaluating language modeling,
    reasoning, and instruction-following capabilities. Comparing Llama 2-7B with
    Llama 3-based models allows us to study the effect of model generation and
    pretraining improvements on downstream benchmark performance.

    \item \textbf{RoBERTa-large.}
    RoBERTa-large~\cite{liu2019roberta} is an encoder-only Transformer model based on the BERT
    architecture, pretrained with a robustly optimized masked language modeling
    objective. Unlike the Llama models, RoBERTa-large is primarily designed for
    natural language understanding tasks rather than autoregressive text
    generation. It is commonly used for classification, entailment, reading
    comprehension, and semantic matching tasks. We include RoBERTa-large as a
    representative discriminative backbone to evaluate the generality of our
    method beyond decoder-only large language models.
\end{itemize}

\section{Baselines}
\label{app:baseline}
We compare our method with several representative efficient fine-tuning
baselines, including LoRA~\cite{hu2022lora}, LOMO~\cite{lv2024full}, GaLore~\cite{zhaogalore}, BAdam~\cite{luo2024badam}, HiFT~\cite{liu2024hift}, and APOLLO~\cite{zhu2025apollo}. These methods cover both parameter-efficient fine-tuning and memory-efficient full-parameter optimization strategies, providing a comprehensive comparison across different training paradigms.

\begin{itemize}[leftmargin=*]
    \item \textbf{LoRA.}
    LoRA~\cite{hu2022lora}, short for Low-Rank Adaptation, is a widely used parameter-efficient
    fine-tuning method. Instead of updating all parameters of the pre-trained
    model, LoRA freezes the original model weights and injects trainable
    low-rank matrices into selected linear layers. During fine-tuning, only
    these additional low-rank parameters are optimized, which significantly
    reduces the number of trainable parameters and optimizer states. Since the
    learned LoRA weights can be merged into the original model weights after
    training, LoRA introduces little or no additional inference latency.

    \item \textbf{LOMO.}
    LOMO~\cite{lv2024full}, or LOw-Memory Optimization, is a memory-efficient optimization method
    designed for full-parameter fine-tuning of large language models. Unlike
    parameter-efficient methods that update only a small subset of parameters,
    LOMO aims to update the full model while reducing memory consumption. It
    fuses gradient computation and parameter update into a single step, thereby
    avoiding the storage of large intermediate gradients and optimizer states.
    This makes full-parameter fine-tuning more feasible under limited GPU
    memory budgets.

    \item \textbf{GaLore.}
    GaLore~\cite{zhaogalore}, short for Gradient Low-Rank Projection, is a memory-efficient
    training strategy that performs full-parameter learning through low-rank
    gradient projection. Instead of directly storing and updating full-rank
    optimizer states, GaLore projects gradients into a low-rank subspace and
    applies optimization in the compressed space. This design reduces optimizer
    memory while still allowing all model parameters to be updated, making it a
    competitive alternative to both standard full fine-tuning and
    parameter-efficient fine-tuning methods.

    \item \textbf{BAdam.}
    BAdam~\cite{luo2024badam} is a memory-efficient full-parameter optimization method based on
    block coordinate descent with Adam-style updates. Rather than updating all
    parameters simultaneously, BAdam divides the model into blocks and updates
    only a subset of parameters at each optimization stage. By maintaining
    optimizer states only for the currently active block, BAdam substantially
    reduces GPU memory usage while preserving the benefits of adaptive
    optimization. It is particularly suitable for fine-tuning large language
    models when full Adam-based optimization is too memory-intensive.

    \item \textbf{HiFT.}
    HiFT~\cite{liu2024hift}, or Hierarchical Full-parameter Fine-Tuning, is an efficient
    full-parameter fine-tuning strategy that updates only a subset of model
    parameters at each training step. Across multiple steps, different
    parameter subsets are selected so that the full model can eventually be
    adapted. This hierarchical update mechanism reduces the amount of gradients
    and optimizer states that need to reside in GPU memory at any given time,
    while avoiding the introduction of additional trainable modules or
    inference-time overhead.

    \item \textbf{APOLLO.}
    APOLLO~\cite{zhu2025apollo}, short for Approximated Gradient Scaling for Memory-Efficient LLM
    Optimization, is a memory-efficient optimizer designed to achieve
    AdamW-level performance with memory consumption closer to SGD. APOLLO
    approximates adaptive learning-rate scaling using auxiliary low-rank
    optimizer states based on random projection, thereby greatly reducing the
    memory overhead associated with Adam-style second-order statistics. It can
    be applied to both large-scale pre-training and full-parameter fine-tuning,
    and serves as a strong baseline for memory-efficient optimization of large
    language models.
\end{itemize}

\section{Implementation Details}
\label{app:implement}

For the baseline methods, including GaLore, BAdam, and APOLLO, we implement all experiments using the LLaMA-Factory framework~\cite{zheng2024llamafactory}. RoBERTa-large experiments are conducted on a server equipped with NVIDIA RTX 4090 GPUs, while all other methods are evaluated on NVIDIA H800-80GB GPUs. All reported experimental results are averaged over five random seeds: 7, 42, 123, 1234, and 12345.

To ensure a fair comparison of memory efficiency, we report peak GPU memory usage for all methods. The memory comparison is mainly intended to demonstrate the memory advantage of our proposed method. For APOLLO, we only compare the memory usage under the rank-1 setting. For ChunkFT, we empirically recommend setting the hyperparameters $K$ and $T$ to the same value. All experiments are conducted with gradient checkpointing enabled, while DeepSpeed was not used. Peak GPU memory consumption was measured using the torch.cuda.max\_memory\_allocated() function. Other hyperparameters and training configurations are provided in Tables~\ref{tab:hyperparameter1},~\ref{tab:hyperparameter3}, and~\ref{tab:hyperparameter2}.

\begin{table*}[ht]
\captionsetup{skip=2pt}
\caption{Hyperparameters of SuperGLUE tasks and instruction tuning.}
\label{tab:hyperparameter1}
\begin{adjustbox}{max width=0.5\textwidth, center}
\includegraphics[width=\textwidth]{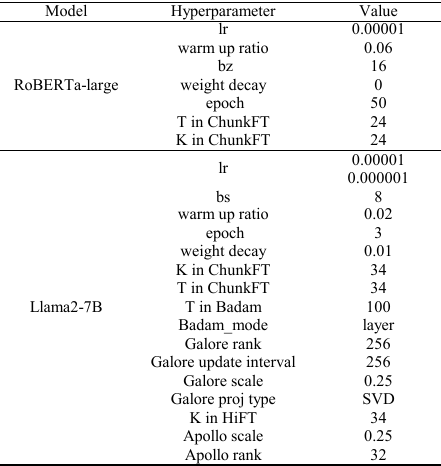}
\end{adjustbox}
\vspace{-2.0em}
\end{table*}

\begin{table*}[ht]
\captionsetup{skip=2pt}
\caption{Hyperparameters of instruction tuning.}
\label{tab:hyperparameter3}
\begin{adjustbox}{max width=0.5\textwidth, center}
\includegraphics[width=\textwidth]{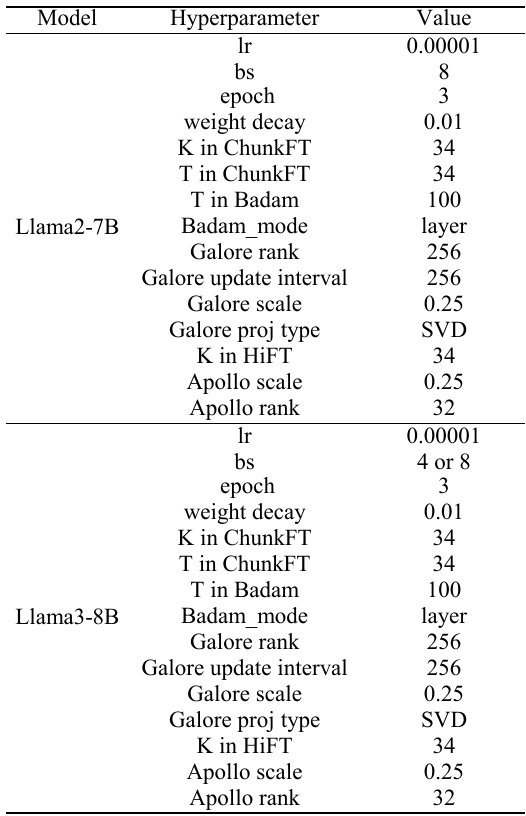}
\end{adjustbox}
\vspace{-2.0em}
\end{table*}

\begin{table*}[ht]
\captionsetup{skip=2pt}
\caption{Hyperparameters of math fine-tuning.}
\label{tab:hyperparameter2}
\begin{adjustbox}{max width=0.5\textwidth, center}
\includegraphics[width=\textwidth]{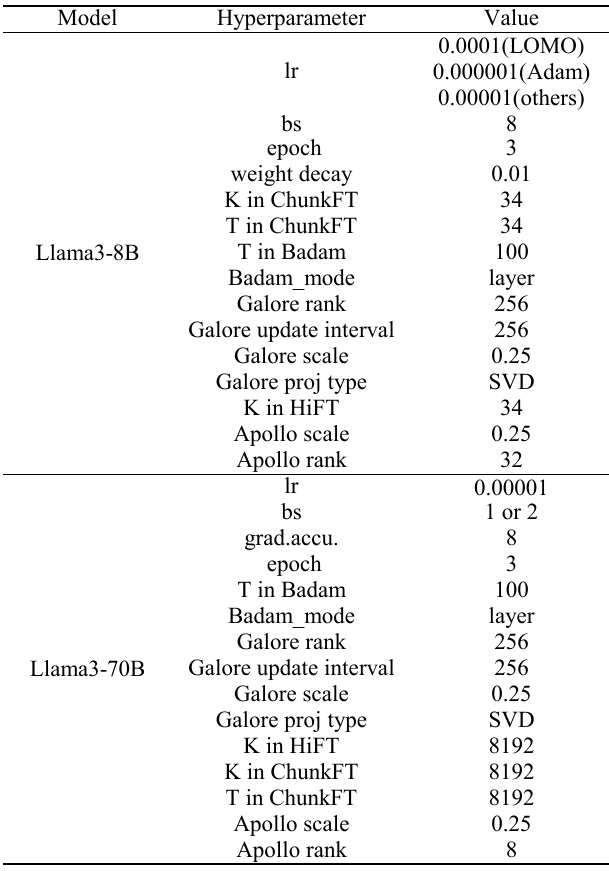}
\end{adjustbox}
\vspace{-2.0em}
\end{table*}

\section{BP Time Comparison}
\label{app:bp_time_comparison}

We compare methods using the same metric: the gradient-generation backward computation required to cover all trainable parameters once. 
Let \(C\) denote the cost of generating dense gradients for all trainable parameters once.


\paragraph{Adam.}
Adam~\cite{loshchilov2017decoupled} generates dense gradients for all trainable parameters in one step. 
Thus,
\begin{equation}
C_{\mathrm{Adam}} = C,
\qquad
\frac{C_{\mathrm{Adam}}}{C}=1.
\end{equation}

\paragraph{LOMO.}
LOMO~\cite{lv2024full} reduces memory by fusing gradient computation and parameter update. 
However, it still needs full-parameter gradient information to update all parameters once. 
Thus,
\begin{equation}
C_{\mathrm{LOMO}}\approx C,
\qquad
\frac{C_{\mathrm{LOMO}}}{C}\approx 1.
\end{equation}

\paragraph{LoRA.}
LoRA~\cite{hu2022lora} only generates gradients for low-rank adapter parameters. 
For a square weight matrix of dimension \(m\times m\) and rank \(r\), its gradient-generation cost is
\begin{equation}
C_{\mathrm{LoRA}}
=
\mathcal O\left(\frac{r}{m}\right)C.
\end{equation}
Hence,
\begin{equation}
\frac{C_{\mathrm{LoRA}}}{C}
=
\mathcal O\left(\frac{r}{m}\right).
\end{equation}
This cost is not a full-parameter optimization cost because the pretrained weights are frozen.

\paragraph{BAdam.}
BAdam~\cite{luo2024badam} performs block-wise rotating optimization. 
Consider the case where the blocks are ordered by network depth, and at each step only one block is selected for update while the other blocks are frozen. 
Although parameter gradients are only materialized for the selected block, backpropagation must still traverse the suffix computation graph from the loss to the selected block in order to compute its gradient.

Let \(C_{\ge k}\) denote the backward gradient-generation cost of the suffix subgraph from block \(k\) to the output. 
Then the cost of updating block \(k\) is
\begin{equation}
C_{\mathrm{BAdam}}(k)\approx C_{\ge k}.
\end{equation}
A full cycle over all \(K\) depth-ordered blocks gives
\begin{equation}
C_{\mathrm{BAdam}}
\approx
\sum_{k=1}^{K} C_{\ge k}.
\end{equation}
If the blocks have approximately uniform backward cost, then
\begin{equation}
C_{\ge k}\approx \frac{K-k+1}{K}C,
\end{equation}
and hence
\begin{equation}
C_{\mathrm{BAdam}}
\approx
\sum_{k=1}^{K}\frac{K-k+1}{K}C
=
\frac{K+1}{2}C.
\end{equation}
Therefore,
\begin{equation}
\frac{C_{\mathrm{BAdam}}}{C}
\approx
\frac{K+1}{2}.
\end{equation}
Thus, under depth-wise block rotation, BAdam incurs repeated suffix backward computation across a full cycle. 
Its cost scales with the number of blocks, but more precisely as \((K+1)/2\) under the uniform-cost assumption, rather than \(K\).

\paragraph{HiFT.}
HiFT~\cite{liu2024hift} performs layer-wise rotating updates. 
At each step, only one layer or block is selected for optimization, while the remaining layers are frozen. 
However, freezing non-selected weights does not localize backpropagation to the selected layer alone. 
To compute the gradient of a selected intermediate layer, the loss gradient must still be propagated backward through all subsequent layers.

Let \(C_{\ge k}\) denote the backward gradient-generation cost from layer or block \(k\) to the output. 
Then the per-step cost is
\begin{equation}
C_{\mathrm{HiFT}}(k)\approx C_{\ge k}.
\end{equation}
A full rotation over \(K\) depth-ordered layers or blocks gives
\begin{equation}
C_{\mathrm{HiFT}}
\approx
\sum_{k=1}^{K} C_{\ge k}.
\end{equation}
For approximately uniform layer or block costs,
\begin{equation}
C_{\ge k}\approx \frac{K-k+1}{K}C.
\end{equation}
Therefore,
\begin{equation}
C_{\mathrm{HiFT}}
\approx
\sum_{k=1}^{K}\frac{K-k+1}{K}C
=
\frac{K+1}{2}C,
\end{equation}
and
\begin{equation}
\frac{C_{\mathrm{HiFT}}}{C}
\approx
\frac{K+1}{2}.
\end{equation}
Thus, HiFT reduces parameter-gradient materialization to the selected layer, but it still incurs suffix backward computation. 
Across a full depth-wise rotation, this leads to a cost larger than a single full backward pass and approximately \((K+1)/2\) times \(C\) under uniform layer costs.

\paragraph{\textsc{ChunkFT}.}
\textsc{ChunkFT} applies chunk selection before gradient materialization. 
Let \(C_k\) be the cost of generating gradients for chunk \(\mathcal C_k\). 
Since the chunks form a disjoint partition,
\begin{equation}
\sum_{k=1}^{K} C_k = C.
\end{equation}
Thus,
\begin{equation}
C_{\textsc{ChunkFT}}
=
\sum_{k=1}^{K} C_k
=
C,
\qquad
\frac{C_{\textsc{ChunkFT}}}{C}=1.
\end{equation}

\paragraph{Summary.}
\begin{equation}
\begin{array}{c|c}
\text{Method} & \text{Grad.-generation BP per full-param. cycle} \\
\hline
\text{Adam} & 1\times \\
\text{LOMO} & \approx 1\times \\
\text{LoRA} & \mathcal O(r/m) \\
\text{BAdam} & \approx \sum_{k=1}^{K}C_{\ge k}/C
\approx (K+1)/2 \\
\text{HiFT} & \approx \sum_{k=1}^{K}C_{\ge k}/C
\approx (K+1)/2 \\
\textsc{ChunkFT} & \approx 1\times \\
\hline
\end{array}
\end{equation}

The difference comes from where sparsity is applied. 
BAdam and HiFT select blocks after dense or near-dense gradient generation, whereas \textsc{ChunkFT} selects chunks before gradient materialization.

\section{Analysis of Memory Jitter}
\label{app:memory_jitter}

Besides peak memory, we also analyze memory jitter, which measures the variation of GPU memory across training steps. 
For a training method with step-wise GPU memory $\mathcal M_t$, we define the memory jitter ratio as
\begin{equation}
\mathrm{Jitter}
=
\frac{
\max_t \mathcal M_t - \min_t \mathcal M_t
}{
\frac{1}{T}\sum_{t=1}^{T}\mathcal M_t
}.
\end{equation}
A smaller value indicates a more stable memory footprint.

For dense methods such as AdamW, LOMO, and LoRA, the training pattern is fixed across steps, so the algorithm-induced memory jitter is small:
\begin{equation}
\mathrm{Jitter}_{\mathrm{Adam/LOMO/LoRA}}
\approx 0.
\end{equation}

For layer-wise rotating methods such as BAdam and HiFT, the active partition is tied to model layers or modules. 
Let $M_k^{\mathrm{layer}}$ denote the number of trainable parameters in the $k$-th layer-wise partition. 
The step-wise memory can be approximated as
\begin{equation}
\mathcal M_k^{\mathrm{layer}}
=
2M
+
16M_k^{\mathrm{layer}}
+
\mathcal A_k
+
\Delta_k,
\end{equation}
where $\mathcal A_k$ denotes activation memory and $\Delta_k$ denotes temporary buffers. 
Since layer-wise partitions can be highly imbalanced, especially for embeddings, MLPs, and output heads, the memory jitter becomes
\begin{equation}
\mathrm{Jitter}_{\mathrm{layer}}
=
\frac{
\max_k \mathcal M_k^{\mathrm{layer}}
-
\min_k \mathcal M_k^{\mathrm{layer}}
}{
\frac{1}{K}
\sum_{k=1}^{K}
\mathcal M_k^{\mathrm{layer}}
}.
\end{equation}
This value can be large when one layer-wise partition contains significantly more parameters or optimizer states than others.

In contrast, \textsc{ChunkFT} constructs chunks according to byte-level training cost rather than layer identity. 
For each chunk $\mathcal C_k$, we estimate
\begin{equation}
B(\mathcal C_k)
=
B_{\mathrm{param}}(\mathcal C_k)
+
B_{\mathrm{grad}}(\mathcal C_k)
+
B_{\mathrm{master}}(\mathcal C_k)
+
B_{\mathrm{opt}}(\mathcal C_k),
\end{equation}
and partition parameters so that
\begin{equation}
B(\mathcal C_k)
\approx
\frac{1}{K}
\sum_{j=0}^{K-1}B(\mathcal C_j).
\end{equation}
The step-wise memory of \textsc{ChunkFT} is therefore
\begin{equation}
\mathcal M_k^{\textsc{ChunkFT}}
=
2M
+
B(\mathcal C_k)
+
\mathcal A_k
+
\Delta_k.
\end{equation}
Ignoring minor variations from activations and temporary buffers, byte-balanced chunking yields
\begin{equation}
\max_k \mathcal M_k^{\textsc{ChunkFT}}
-
\min_k \mathcal M_k^{\textsc{ChunkFT}}
\approx 0.
\end{equation}
More generally, if the chunking imbalance is bounded by
\begin{equation}
\epsilon_B
=
\max_k
\left|
B(\mathcal C_k)
-
\frac{1}{K}
\sum_{j=0}^{K-1}
B(\mathcal C_j)
\right|,
\end{equation}
then
\begin{equation}
\mathrm{Jitter}_{\textsc{ChunkFT}}
\lesssim
\frac{
2\epsilon_B
}{
2M+\frac{1}{K}\sum_{j=0}^{K-1}B(\mathcal C_j)
}.
\end{equation}
Thus, unlike layer-wise methods whose memory jitter is governed by layer-size imbalance, \textsc{ChunkFT} explicitly controls memory jitter through byte-balanced chunk construction.

\section{Convergence Analysis of \textsc{ChunkFT}}
\label{app:chunkft_convergence}

In this section, we provide a convergence analysis of \textsc{ChunkFT}. 
The analysis follows the standard non-convex block-coordinate optimization framework and shows that rotating chunk updates converge to a first-order stationary point under smoothness and bounded partial-derivative assumptions.

\subsection{Notation}

Let the trainable parameters be partitioned into $K$ disjoint chunks:
\begin{equation}
\theta =
(\theta_1,\theta_2,\ldots,\theta_K),
\end{equation}
where $\theta_i$ denotes the parameters in the $i$-th chunk. 
Let $\mathcal L(\theta)$ denote the training objective, which is assumed to be lower bounded by $\mathcal L^\star$:
\begin{equation}
\mathcal L(\theta)\geq \mathcal L^\star,\qquad \forall \theta .
\end{equation}

\textsc{ChunkFT} updates chunks in a rotating manner. 
We use $t=0,1,\ldots,T-1$ to index chunk epochs, $i=1,\ldots,K$ to index chunks, and $h=0,\ldots,H-1$ to index the inner updates performed on a selected chunk. 
At chunk epoch $t$, the active chunk is
\begin{equation}
i_t = (t \bmod K) + 1.
\end{equation}
For this active chunk, \textsc{ChunkFT} performs $H$ inner updates before switching to the next chunk. 
We denote the model parameter after the $h$-th inner update of chunk epoch $t$ by $\theta^{t,h}$, with
\begin{equation}
\theta^{t,0}
\end{equation}
being the parameter at the beginning of chunk epoch $t$. 
The partial derivative with respect to the active chunk is denoted by
\begin{equation}
g_{i_t}^{t,h}
=
\nabla_{i_t}\mathcal L(\theta^{t,h}).
\end{equation}

The update rule is
\begin{equation}
\theta_{i_t}^{t,h+1}
=
\theta_{i_t}^{t,h}
-
\eta g_{i_t}^{t,h},
\end{equation}
and all inactive chunks remain unchanged:
\begin{equation}
\theta_j^{t,h+1}
=
\theta_j^{t,h},
\qquad j\neq i_t .
\end{equation}
After $H$ inner updates, we set
\begin{equation}
\theta^{t+1,0}=\theta^{t,H}.
\end{equation}

For notational simplicity, the proof is written for the block-gradient version of \textsc{ChunkFT}. 
The AdamW implementation used in practice can be viewed as a chunk-wise preconditioned variant of this update; we discuss this at the end of the section.

\subsection{Assumptions}

We make the following two assumptions. Assumption D.1 is standard for analyzing block
descent-type methods~\cite{wright2015coordinate}. Assumption D.2 is commonly used in the analysis of Adam~\cite{defossez2020simple}. We adopt this assumption for simplicity of presentation, noting that it can be provably ensured~\cite{li2023convergence}.

\paragraph{Assumption D.1. Smoothness.}
The loss function $\mathcal L$ is $L$-smooth. 
When restricted to the $i$-th chunk, it is $L_i$-smooth. 
Mathematically, for any $\theta^1,\theta^2$,
\begin{equation}
\left\|
\nabla \mathcal L(\theta^1)
-
\nabla \mathcal L(\theta^2)
\right\|
\leq
L
\left\|
\theta^1-\theta^2
\right\|,
\end{equation}
and for each chunk $i=1,\ldots,K$,
\begin{equation}
\left\|
\nabla_i \mathcal L(\theta^1)
-
\nabla_i \mathcal L(\theta^2)
\right\|
\leq
L_i
\left\|
\theta_i^1-\theta_i^2
\right\|.
\end{equation}
Equivalently, for an update only on chunk $i$,
\begin{equation}
\mathcal L(\theta+U_i d_i)
\leq
\mathcal L(\theta)
+
\left\langle
\nabla_i\mathcal L(\theta),
d_i
\right\rangle
+
\frac{L_i}{2}
\|d_i\|^2,
\end{equation}
where $U_i d_i$ inserts $d_i$ into the $i$-th chunk and leaves other chunks unchanged.

We define
\begin{equation}
\overline L = \max_{i=1,\ldots,K}L_i,
\qquad
\underline L = \min_{i=1,\ldots,K}L_i.
\end{equation}

\paragraph{Assumption D.2. Bounded partial derivatives.}
\textsc{ChunkFT} has bounded partial derivatives along its optimization trajectory. 
That is, there exists $G>0$ such that
\begin{equation}
\left\|
g_{i_t}^{t,h}
\right\|
\leq
G,
\qquad
\forall t,\ h .
\end{equation}
Here
\begin{equation}
g_{i_t}^{t,h}
=
\nabla_{i_t}\mathcal L(\theta^{t,h})
\end{equation}
is the partial derivative with respect to the active chunk at chunk epoch $t$ and inner update $h$.

\subsection{One-Step Descent}

For one inner update on active chunk $i_t$, we have
\begin{equation}
\theta^{t,h+1}
=
\theta^{t,h}
-
\eta U_{i_t}g_{i_t}^{t,h}.
\end{equation}
By block-wise smoothness in Assumption D.1,
\begin{equation}
\mathcal L(\theta^{t,h+1})
\leq
\mathcal L(\theta^{t,h})
-
\eta
\left\|
g_{i_t}^{t,h}
\right\|^2
+
\frac{L_{i_t}\eta^2}{2}
\left\|
g_{i_t}^{t,h}
\right\|^2.
\end{equation}
Thus,
\begin{equation}
\mathcal L(\theta^{t,h+1})
\leq
\mathcal L(\theta^{t,h})
-
\eta
\left(
1-\frac{L_{i_t}\eta}{2}
\right)
\left\|
g_{i_t}^{t,h}
\right\|^2.
\end{equation}
If
\begin{equation}
0<\eta\leq \frac{1}{\overline L},
\end{equation}
then
\begin{equation}
1-\frac{L_{i_t}\eta}{2}
\geq
\frac{1}{2}.
\end{equation}
Therefore,
\begin{equation}
\mathcal L(\theta^{t,h+1})
\leq
\mathcal L(\theta^{t,h})
-
\frac{\eta}{2}
\left\|
g_{i_t}^{t,h}
\right\|^2.
\end{equation}
Equivalently,
\begin{equation}
\left\|
g_{i_t}^{t,h}
\right\|^2
\leq
\frac{2}{\eta}
\left(
\mathcal L(\theta^{t,h})
-
\mathcal L(\theta^{t,h+1})
\right).
\end{equation}

\subsection{Average Active-Chunk Stationarity}

Summing the one-step descent inequality over all chunk epochs $t=0,\ldots,T-1$ and inner updates $h=0,\ldots,H-1$, we obtain
\begin{equation}
\frac{\eta}{2}
\sum_{t=0}^{T-1}
\sum_{h=0}^{H-1}
\left\|
g_{i_t}^{t,h}
\right\|^2
\leq
\mathcal L(\theta^{0,0})
-
\mathcal L(\theta^{T,0}).
\end{equation}
Since $\mathcal L$ is lower bounded by $\mathcal L^\star$,
\begin{equation}
\frac{\eta}{2}
\sum_{t=0}^{T-1}
\sum_{h=0}^{H-1}
\left\|
g_{i_t}^{t,h}
\right\|^2
\leq
\mathcal L(\theta^{0,0})
-
\mathcal L^\star.
\end{equation}
Therefore,
\begin{equation}
\frac{1}{TH}
\sum_{t=0}^{T-1}
\sum_{h=0}^{H-1}
\left\|
g_{i_t}^{t,h}
\right\|^2
\leq
\frac{
2\left(
\mathcal L(\theta^{0,0})-\mathcal L^\star
\right)
}{
\eta TH
}.
\end{equation}
This implies
\begin{equation}
\min_{0\leq t<T,\ 0\leq h<H}
\left\|
g_{i_t}^{t,h}
\right\|^2
\leq
\frac{
2\left(
\mathcal L(\theta^{0,0})-\mathcal L^\star
\right)
}{
\eta TH
}.
\end{equation}
Hence, as the number of training updates increases, the active-chunk gradient converges to zero in the ergodic sense:
\begin{equation}
\frac{1}{TH}
\sum_{t=0}^{T-1}
\sum_{h=0}^{H-1}
\left\|
\nabla_{i_t}\mathcal L(\theta^{t,h})
\right\|^2
=
\mathcal O\left(\frac{1}{TH}\right).
\end{equation}

\subsection{Full-Gradient Stationarity over Completed Rotations}

The preceding result controls the gradient of the active chunk at its update time. 
We now relate it to the full gradient at the beginning of completed rotations.

Let
\begin{equation}
\tau_r = rK
\end{equation}
denote the beginning chunk epoch of the $r$-th full rotation, and let
\begin{equation}
R = \left\lfloor \frac{T}{K} \right\rfloor
\end{equation}
be the number of completed rotations. 
Within each rotation, every chunk is activated once. 
For each chunk $i$, let
\begin{equation}
t_{r,i}\in \{\tau_r,\tau_r+1,\ldots,\tau_r+K-1\}
\end{equation}
be the chunk epoch in which chunk $i$ is activated during rotation $r$.

By smoothness,
\begin{equation}
\left\|
\nabla_i\mathcal L(\theta^{\tau_r,0})
-
\nabla_i\mathcal L(\theta^{t_{r,i},0})
\right\|
\leq
L_i
\left\|
\theta^{\tau_r,0}
-
\theta^{t_{r,i},0}
\right\|.
\end{equation}
Using
\begin{equation}
\|a\|^2
\leq
2\|b\|^2+2\|a-b\|^2,
\end{equation}
we have
\begin{equation}
\left\|
\nabla_i\mathcal L(\theta^{\tau_r,0})
\right\|^2
\leq
2
\left\|
\nabla_i\mathcal L(\theta^{t_{r,i},0})
\right\|^2
+
2L_i^2
\left\|
\theta^{\tau_r,0}
-
\theta^{t_{r,i},0}
\right\|^2.
\end{equation}
Summing over all chunks gives
\begin{equation}
\left\|
\nabla\mathcal L(\theta^{\tau_r,0})
\right\|^2
\leq
2
\sum_{i=1}^{K}
\left\|
\nabla_i\mathcal L(\theta^{t_{r,i},0})
\right\|^2
+
2\overline L^2
\sum_{i=1}^{K}
\left\|
\theta^{\tau_r,0}
-
\theta^{t_{r,i},0}
\right\|^2.
\end{equation}

We bound the movement term. 
During one rotation, at most $KH$ inner updates are performed. 
By Assumption D.2, each update satisfies
\begin{equation}
\left\|
\theta^{t,h+1}
-
\theta^{t,h}
\right\|
=
\eta
\left\|
g_{i_t}^{t,h}
\right\|
\leq
\eta G.
\end{equation}
Therefore, for any $t_{r,i}$ in the same rotation,
\begin{equation}
\left\|
\theta^{\tau_r,0}
-
\theta^{t_{r,i},0}
\right\|
\leq
KH\eta G,
\end{equation}
and hence
\begin{equation}
\left\|
\theta^{\tau_r,0}
-
\theta^{t_{r,i},0}
\right\|^2
\leq
K^2H^2\eta^2G^2.
\end{equation}
Thus,
\begin{equation}
\left\|
\nabla\mathcal L(\theta^{\tau_r,0})
\right\|^2
\leq
2
\sum_{i=1}^{K}
\left\|
\nabla_i\mathcal L(\theta^{t_{r,i},0})
\right\|^2
+
2K^3H^2\overline L^2\eta^2G^2.
\end{equation}

Averaging over $R$ completed rotations, we obtain
\begin{equation}
\frac{1}{R}
\sum_{r=0}^{R-1}
\left\|
\nabla\mathcal L(\theta^{\tau_r,0})
\right\|^2
\leq
\frac{2}{R}
\sum_{r=0}^{R-1}
\sum_{i=1}^{K}
\left\|
\nabla_i\mathcal L(\theta^{t_{r,i},0})
\right\|^2
+
2K^3H^2\overline L^2\eta^2G^2.
\end{equation}

Since $R=\lfloor T/K\rfloor$, the first term is controlled by the active-chunk stationarity bound:
\begin{equation}
\frac{1}{R}
\sum_{r=0}^{R-1}
\sum_{i=1}^{K}
\left\|
\nabla_i\mathcal L(\theta^{t_{r,i},0})
\right\|^2
=
\mathcal O\left(\frac{K}{\eta TH}\right).
\end{equation}
Therefore,
\begin{equation}
\frac{1}{R}
\sum_{r=0}^{R-1}
\left\|
\nabla\mathcal L(\theta^{\tau_r,0})
\right\|^2
=
\mathcal O\left(\frac{K}{\eta TH}\right)
+
\mathcal O\left(K^3H^2\eta^2\right).
\end{equation}

Choosing
\begin{equation}
\eta =
\mathcal O
\left(
\frac{1}{K^{2/3}H\,T^{1/3}}
\right)
\end{equation}
balances the two terms and yields
\begin{equation}
\frac{1}{R}
\sum_{r=0}^{R-1}
\left\|
\nabla\mathcal L(\theta^{\tau_r,0})
\right\|^2
=
\mathcal O
\left(
\frac{K^{5/3}}{T^{2/3}}
\right),
\end{equation}
for fixed $H$. 
In particular, for fixed $K$ and $H$,
\begin{equation}
\liminf_{r\to\infty}
\left\|
\nabla\mathcal L(\theta^{\tau_r,0})
\right\|^2
=
0.
\end{equation}

\subsection{Theorem}

\begin{theorem}[Convergence of \textsc{ChunkFT}]
\label{thm:chunkft_convergence}
Suppose Assumptions D.1--D.2 hold and $\mathcal L$ is lower bounded by $\mathcal L^\star$. 
Let \textsc{ChunkFT} update chunks according to the rotating schedule
\begin{equation}
i_t=(t\bmod K)+1,
\end{equation}
and perform $H$ inner updates on each selected chunk. 
If the learning rate satisfies $0<\eta\leq 1/\overline L$, then
\begin{equation}
\frac{1}{TH}
\sum_{t=0}^{T-1}
\sum_{h=0}^{H-1}
\left\|
\nabla_{i_t}\mathcal L(\theta^{t,h})
\right\|^2
\leq
\frac{
2\left(
\mathcal L(\theta^{0,0})-\mathcal L^\star
\right)
}{
\eta TH
}.
\end{equation}
Furthermore, over completed rotations,
\begin{equation}
\frac{1}{R}
\sum_{r=0}^{R-1}
\left\|
\nabla\mathcal L(\theta^{\tau_r,0})
\right\|^2
=
\mathcal O\left(\frac{K}{\eta TH}\right)
+
\mathcal O\left(K^3H^2\eta^2\right),
\end{equation}
where $\tau_r=rK$ and $R=\lfloor T/K\rfloor$. 
With an appropriate diminishing learning rate, for fixed $K$ and $H$, \textsc{ChunkFT} converges to a first-order stationary point:
\begin{equation}
\liminf_{r\to\infty}
\left\|
\nabla\mathcal L(\theta^{\tau_r,0})
\right\|^2
=
0.
\end{equation}
\end{theorem}

\subsection{Remark on AdamW and CPU Offloading}

The proof above is written for block-gradient updates to isolate the effect of rotating chunk optimization. 
In practice, \textsc{ChunkFT} uses AdamW with chunk-wise first- and second-moment states. 
The AdamW update can be written as a preconditioned block update:
\begin{equation}
\theta_{i_t}^{t,h+1}
=
\theta_{i_t}^{t,h}
-
\eta
P_{t,h}^{(i_t)}
g_{i_t}^{t,h}
-
\eta\lambda\theta_{i_t}^{t,h},
\end{equation}
where $P_{t,h}^{(i_t)}$ is a diagonal adaptive preconditioner. 
If the preconditioner is uniformly bounded and positive, i.e.,
\begin{equation}
0<p_{\min}I
\preceq
P_{t,h}^{(i_t)}
\preceq
p_{\max}I,
\end{equation}
the same proof applies up to constants depending on $p_{\min}$ and $p_{\max}$.

CPU offloading does not affect the mathematical update. 
The fp32 master weights and optimizer states of each chunk are stored on CPU and reloaded when the corresponding chunk becomes active. 
Therefore, offloading changes memory residency and data movement, but not the optimization trajectory defined by the chunk-wise update rule.

\subsection{Remark on Stochastic Gradients}

For mini-batch training, the same analysis can be extended by replacing the deterministic partial derivative with a stochastic estimator $g_{i_t}^{t,h}$ satisfying
\begin{equation}
\mathbb E[g_{i_t}^{t,h}\mid \theta^{t,h}]
=
\nabla_{i_t}\mathcal L(\theta^{t,h}),
\qquad
\mathbb E\|g_{i_t}^{t,h}\|^2
\leq
G^2.
\end{equation}
The resulting convergence bound has the same form in expectation, with the squared gradient norm replaced by its expected value.

\section{Limitations}
\label{app:limitations}

\textsc{ChunkFT} reduces GPU memory by activating and optimizing only one byte-balanced parameter chunk at a time, but this design also introduces several practical considerations. 
First, \textsc{ChunkFT} relies on chunk-aware backward implementations for memory-dominant modules such as Embedding, Linear, LayerNorm, and RMSNorm. These operators cover the primary components of modern Transformer architectures, enabling broad applicability in standard LLM training pipelines. 
Extending \textsc{ChunkFT} to architectures with highly customized operators may require additional operator-specific implementations.

Second, \textsc{ChunkFT} keeps optimizer states and fp32 master weights on CPU by default and transfers the active chunk between CPU and GPU during training. Consequently, the end-to-end training efficiency can depend on the underlying system configuration, including host-device interconnect bandwidth, CPU memory bandwidth, and the effectiveness of asynchronous prefetching and offloading. In resource-constrained environments, communication overhead may partially limit the achievable runtime improvement.


\newpage
\section*{NeurIPS Paper Checklist}

\begin{enumerate}

\item {\bf Claims}
    \item[] Question: Do the main claims made in the abstract and introduction accurately reflect the paper's contributions and scope?
    \item[] Answer: \answerYes{} 
    \item[] Justification: The abstract and introduction state the main methodological, theoretical, and empirical claims, and these claims are supported by the method description in Section~\ref{sub:method}, the convergence analysis in Appendix~\ref{app:chunkft_convergence}, and the memory, runtime, and downstream evaluation results in Section~\ref{sub:memory-time} and the experiment section.

\item {\bf Limitations}
    \item[] Question: Does the paper discuss the limitations of the work performed by the authors?
    \item[] Answer: \answerYes{} 
    \item[] Justification: The paper includes Limitations section in Appendix~\ref{app:limitations}, discussing the need for chunk-aware backward operators and the dependence on CPU-GPU transfer bandwidth and asynchronous offloading.

\item {\bf Theory assumptions and proofs}
    \item[] Question: For each theoretical result, does the paper provide the full set of assumptions and a complete (and correct) proof?
    \item[] Answer: \answerYes{} 
    \item[] Justification: The paper states the assumptions for the deterministic convergence result in Appendix~\ref{app:chunkft_convergence}, including smoothness and bounded partial derivatives, and provides the descent argument, stationarity results, theorem, and remarks on AdamW and stochastic gradients.

    \item {\bf Experimental result reproducibility}
    \item[] Question: Does the paper fully disclose all the information needed to reproduce the main experimental results of the paper to the extent that it affects the main claims and/or conclusions of the paper (regardless of whether the code and data are provided or not)?
    \item[] Answer: \answerYes{} 
    \item[] Justification: The algorithm is specified in Algorithm~\ref{alg:mrcopt}, and the experimental setup, benchmarks, baselines, random seeds, hardware, memory measurement protocol, and hyperparameters are described in Section~\ref{sub:memory-time} and Appendix~\ref{app:benchmarks}--\ref{app:implement}.

\item {\bf Open access to data and code}
    \item[] Question: Does the paper provide open access to the data and code, with sufficient instructions to faithfully reproduce the main experimental results, as described in supplemental material?
    \item[] Answer: \answerYes{} 
    \item[] Justification: We provide an anonymous code repository.

\item {\bf Experimental setting/details}
    \item[] Question: Does the paper specify all the training and test details (e.g., data splits, hyperparameters, how they were chosen, type of optimizer) necessary to understand the results?
    \item[] Answer: \answerYes{} 
    \item[] Justification: The paper reports the training and evaluation settings, including backbone models, benchmark datasets, baselines, hardware, random seeds, gradient checkpointing, memory measurement with \texttt{torch.cuda.max\_memory\_allocated()}, and hyperparameters in Appendix~\ref{app:implement}.

\item {\bf Experiment statistical significance}
    \item[] Question: Does the paper report error bars suitably and correctly defined or other appropriate information about the statistical significance of the experiments?
    \item[] Answer: \answerYes{} 
    \item[] Justification: Downstream performance tables report variability over five random seeds (Table 4, 6 , 7 and 8).

\item {\bf Experiments compute resources}
    \item[] Question: For each experiment, does the paper provide sufficient information on the computer resources (type of compute workers, memory, time of execution) needed to reproduce the experiments?
    \item[] Answer: \answerYes{} 
    \item[] Justification: The paper reports the main compute resources used for experiments, including RTX 4090-24GB, and H800-80GB GPUs, and provides wall-clock timing for the main Llama 2-7B BoolQ speed comparison in Table~\ref{tab:speed}.
    
\item {\bf Code of ethics}
    \item[] Question: Does the research conducted in the paper conform, in every respect, with the NeurIPS Code of Ethics \url{https://neurips.cc/public/EthicsGuidelines}?
    \item[] Answer: \answerYes{} 
    \item[] Justification: We adhere to all the ethics guidelines.

\item {\bf Broader impacts}
    \item[] Question: Does the paper discuss both potential positive societal impacts and negative societal impacts of the work performed?
    \item[] Answer: \answerNo{} 
    \item[] Justification: The paper focuses on an optimization method and does not include a dedicated broader-impact discussion. 
    
\item {\bf Safeguards}
    \item[] Question: Does the paper describe safeguards that have been put in place for responsible release of data or models that have a high risk for misuse (e.g., pre-trained language models, image generators, or scraped datasets)?
    \item[] Answer: \answerNA{} 
    \item[] Justification: This paper poses no such risks.

\item {\bf Licenses for existing assets}
    \item[] Question: Are the creators or original owners of assets (e.g., code, data, models), used in the paper, properly credited and are the license and terms of use explicitly mentioned and properly respected?
    \item[] Answer: \answerYes{} 
    \item[] Justification: Existing datasets and code are properly cited.

\item {\bf New assets}
    \item[] Question: Are new assets introduced in the paper well documented and is the documentation provided alongside the assets?
    \item[] Answer: \answerNA{} 
    \item[] Justification: Paper does not exist new assets.

\item {\bf Crowdsourcing and research with human subjects}
    \item[] Question: For crowdsourcing experiments and research with human subjects, does the paper include the full text of instructions given to participants and screenshots, if applicable, as well as details about compensation (if any)? 
    \item[] Answer: \answerNA{} 
    \item[] Justification: This paper does not involve human subjects.

\item {\bf Institutional review board (IRB) approvals or equivalent for research with human subjects}
    \item[] Question: Does the paper describe potential risks incurred by study participants, whether such risks were disclosed to the subjects, and whether Institutional Review Board (IRB) approvals (or an equivalent approval/review based on the requirements of your country or institution) were obtained?
    \item[] Answer: \answerNA{} 
    \item[] Justification: This paper does not involve crowdsourcing or human subjects.

\item {\bf Declaration of LLM usage}
    \item[] Question: Does the paper describe the usage of LLMs if it is an important, original, or non-standard component of the core methods in this research? Note that if the LLM is used only for writing, editing, or formatting purposes and does \emph{not} impact the core methodology, scientific rigor, or originality of the research, declaration is not required.
    \item[] Answer: \answerNA{} 
    \item[] Justification: This paper does not involve LLMs in any meaningful capacity.

\end{enumerate}

\end{document}